\documentclass{bmvc2k}

\title{Explainable Visual Anomaly Detection via Concept Bottleneck Models}

\addauthor{Arianna Stropeni}{arianna.stropeni@studenti.unipd.it}{1}
\addauthor{Valentina Zaccaria}{valentina.zaccaria@unipd.it}{1}
\addauthor{Francesco Borsatti}{francesco.borsatti.1@phd.unipd.it}{1}
\addauthor{Davide Dalle Pezze}{davide.dallepezze@unipd.it}{1}
\addauthor{Manuel Barusco}{manuel.barusco@phd.unipd.it}{1}
\addauthor{Gian Antonio Susto}{gianantonio.susto@unipd.it}{1}

\addinstitution{
Department of Information Engineering 
 University of Padova \\
 Padova, Italy
}

\runninghead{Stropeni et al.}{Explainable Visual Anomaly Detection via CBM}

\def\eg{\emph{e.g}\bmvaOneDot}

\newcommand{\ourmethod}{CONVAD}
\newcommand{\ourdataset}{MVTec AD}

\usepackage{orcidlink}

\usepackage[nolist]{acronym}

\usepackage{array}           
\usepackage{multirow}        
\usepackage{makecell}        
\usepackage{colortbl}        
\usepackage{enumitem}
\usepackage{graphicx}
\usepackage{booktabs}
\usepackage{amssymb}
\usepackage{pifont}
\newcommand{\xmark}{\ding{55}}
\usepackage{diagbox}

\usepackage{booktabs}
\usepackage{colortbl}
\usepackage{xcolor}
\usepackage{multirow}
\usepackage{array}
\usepackage{graphicx}

\definecolor{headerblue}{RGB}{52, 86, 139}
\definecolor{headerlightblue}{RGB}{220, 230, 242}
\definecolor{white}{RGB}{255, 255, 255}
\definecolor{headergray}{RGB}{80, 80, 80}
\definecolor{headerlightgray}{RGB}{220, 220, 220}
\definecolor{rowgray}{RGB}{245, 245, 245}

\begin{document}

\begin{acronym}
  \acro{CBM}{Concept Bottleneck Model}
  \acro{VAD}{Visual Anomaly Detection}
  \acro{AD}{anomaly detection}
  \acro{VLM}{Vision Language Model}
  \acro{XAI}{eXplainable Artificial Intelligence}
  \acro{CoT}{Chain-of-Thought}
  \acro{OOD}{out-of-distribution}
  \acro{LLM}{Large Language Model}
\end{acronym}

\maketitle

    \vspace{-7.5mm}
    \begin{abstract}
    In recent years, \ac{VAD} has attracted significant attention due to its high interpretability, as it provides visual explanations by highlighting anomalous regions within an image.
    However, although these visual explanations can be helpful, they lack a direct and semantically meaningful interpretation for users.
    To address this limitation, we propose extending \acp{CBM} to the VAD setting.
    By learning meaningful concepts, the network can provide human-interpretable descriptions of anomalies, offering a novel and more insightful way to explain them.
    Our main contributions are: (i) we introduce a concept-based framework for anomaly explanation by extending \acp{CBM} to the \ac{VAD} setting for the first time; (ii) we evaluate multiple supervision regimes, ranging from supervised to unsupervised anomaly settings; (iii) we propose a dual-branch architecture that combines a \ac{CBM}-based Concept branch for concept-level explanations with a Visual branch for pixel-level anomaly localization, bridging semantic and spatial interpretability. When evaluated across three well-established VAD benchmarks, our approach, Concept-Aware Visual Anomaly Detection (\ourmethod), achieves performance comparable to classic VAD methods (up to 0.97 I-AUROC on MVTec AD) while providing richer, concept-driven explanations. A user study (582 responses) shows that \ourmethod{} explanations are rated effective in 73.5\% of cases (vs.\ 60.7\% for the SOTA VLM-based explainer Anomaly-OV) and misleading in only 3.1\% (vs.\ 11.7\%). Beyond explanation quality, \acp{CBM} enable human-in-the-loop correction: concept-level interventions further boost detection in the challenging unsupervised setting, closing part of the gap to fully supervised performance.
\end{abstract}

\section{Introduction}\label{sec:intro}

\begin{figure}[thbp]
    \centering
    \begin{minipage}[t]{0.55\linewidth}
        \centering
        \includegraphics[width=\linewidth]{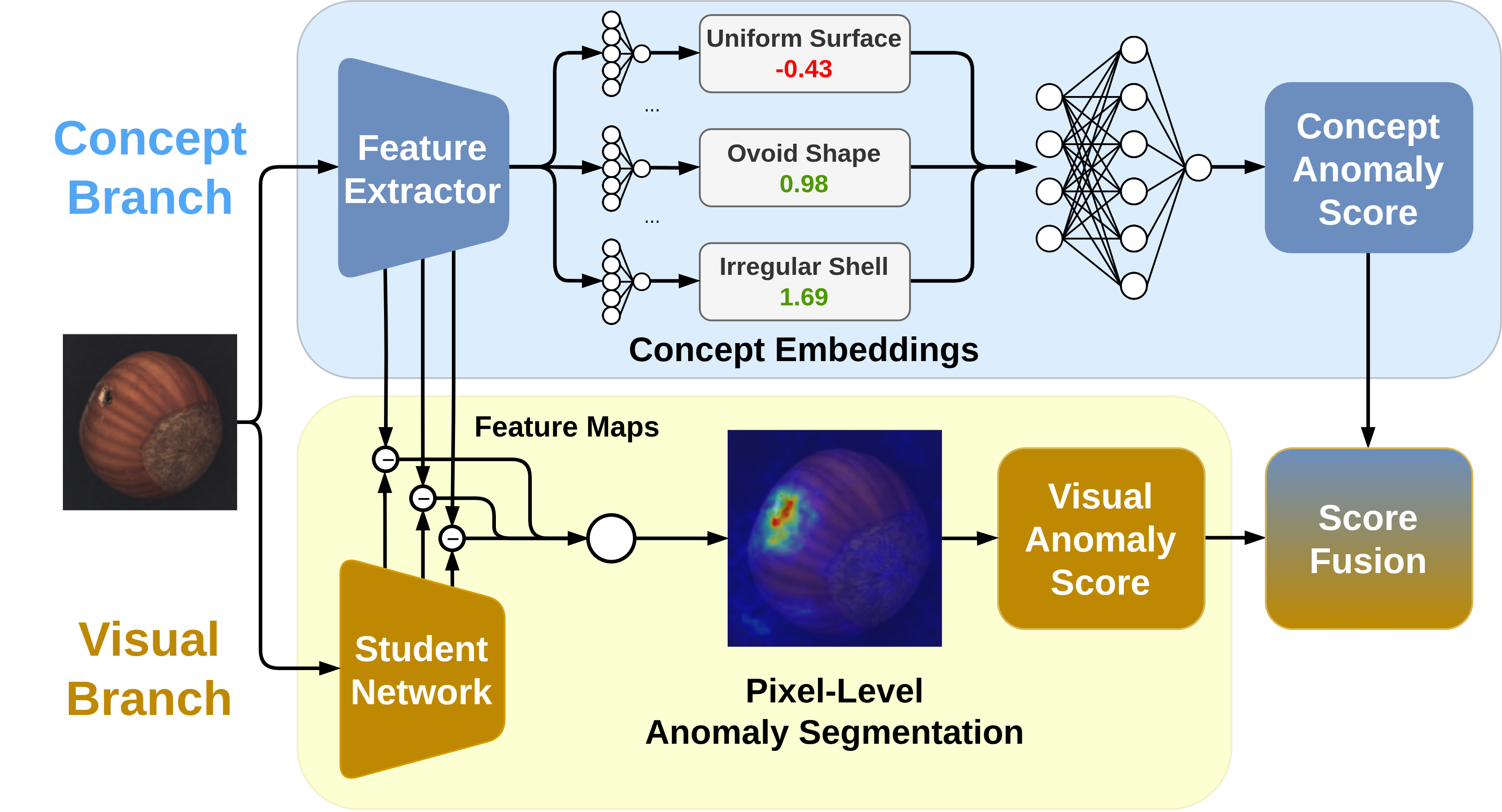}
    \end{minipage}%
    \hspace{5pt}%
    \vrule width 0.3pt%
    \hspace{5pt}%
    \begin{minipage}[t]{0.32\linewidth}
        \centering
        \includegraphics[width=\linewidth]{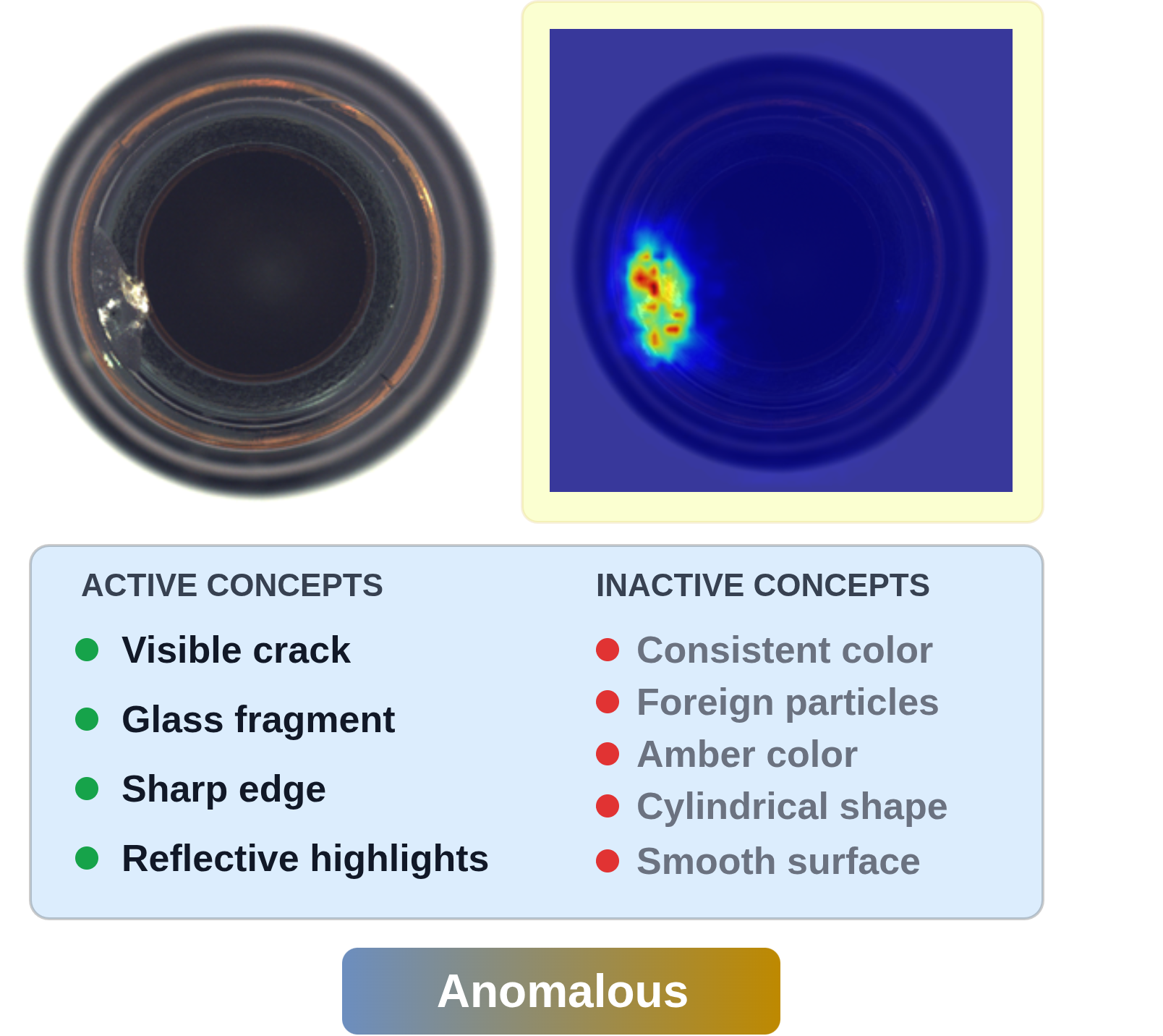}
    \end{minipage}
    \caption{(Left) \ourmethod{} Architecture, (Right) Output example of \ourmethod: the input image, the anomaly heatmap, a list of most active and inactive concepts, and the prediction.}%
    \label{fig:architecture}
\end{figure}

\acl{VAD} (\ac{VAD}) aims to detect anomalies in images while highlighting the pixels responsible for the anomaly.
These capabilities are particularly relevant in domains such as manufacturing, medicine, and surveillance~\cite{mvtec, bmad, chan2021segmentmeifyoucanbenchmarkanomalysegmentation}.

Although \ac{VAD} models can produce anomaly heatmaps, it is questionable whether these visual localizations alone are sufficient for interpretability. Indeed, identifying where an anomaly occurs is sometimes insufficient to understand what is actually anomalous, since this depends on the context and on the underlying notion of normality. 
In this work, we address this limitation by complementing spatial anomaly heatmaps with semantic explanations, called \emph{concepts}, by considering a different \ac{VAD} paradigm based on Concept Bottleneck Models~\cite{koh2020concept} (\acp{CBM}). 
Unlike standard \ac{VAD} approaches, \acp{CBM} rely on supervised concept annotations and explicitly learn intermediate, human-interpretable concepts that drive the final prediction. Incorporating \acp{CBM} into the \ac{VAD} framework enables the system to provide not only image- and pixel-level labels but also a set of interpretable concepts underlying each prediction, which can be used by the end user to understand model outputs and enhance the decision-making process.
In addition, \acp{CBM} allow users to intervene on intermediate concept activations, aiding the model in making correct predictions and enabling seamless \textbf{human-machine collaboration}, which is not possible in standard \ac{VAD} models.

A potential limitation of \acp{CBM} is that they require supervised training. This differs from the traditional \ac{VAD} setting which assumes only normal samples available, since collecting and annotating anomalous examples is expensive, especially in industrial settings. Despite this, a growing body of recent studies has shown that incorporating limited labeled information into \ac{VAD} can substantially improve anomaly detection performance~\cite{sun2025unseen, rolih2025no}. Such supervision may take the form of weak labels or a few real anomalous samples.  Overall, these findings suggest that a balanced trade-off between annotation effort and model capability is both practical and beneficial. Following this direction, our approach adopts a supervised setting under limited supervision, considering scenarios with either very few real anomalies or synthetic anomalies. This minimal annotation effort is compensated by gains in interpretability and more effective human-collaboration.

Our contributions can be summarized as follows:
\begin{enumerate}[label=(\roman*), wide=0pt, itemsep=0.5em]
\item \textbf{Adaptation of \acp{CBM} to \ac{VAD}.}
To the best of our knowledge, this work is the first to successfully adapt \acp{CBM} to the \ac{VAD} scenario. This adaptation is non-trivial and requires addressing several challenges.
\acp{CBM} require a \textit{Concept Dataset} for training, where each sample is annotated with concepts from a predefined vocabulary. 
Since no such dataset exists for \ac{VAD}, to conduct an extensive evaluation of our method, we adopt an automated, VLM-based pipeline for extracting meaningful concepts and annotating industrial images. This is an existing approach for \acp{CBM} literature in other domains~\cite{oikarinen2023label, yang2023language, srivastava2024vlg}.
We publicly release the resulting Concept Datasets for three major industrial \ac{VAD} benchmarks (MVTec AD, VisA, and Real-IAD) to facilitate future research, along with all code (see Supplementary Material) used for the experiments.

\item \textbf{Evaluation of supervision regimes} and synthetic anomaly generation.
We benchmark multiple training settings with varying levels of anomalous data availability: full supervision with all real anomalies, semi-supervision with few real samples, and settings that rely exclusively on synthetically generated defects produced by a dedicated pipeline. This provides a thorough analysis of the performance-labeling effort trade-off, while allowing us to progressively reduce the reliance on real anomalous data and move closer to the standard \ac{VAD} paradigm. For each regime, we compare results with state-of-the-art and standard baselines.

\item \textbf{Dual-branch architecture} for concept-level and pixel-level explanations.
\ac{VAD} approaches provide pixel-level anomaly localization, a capability that standard \acp{CBM} lack since they are limited to sample-level predictions. We propose a novel two-branch architecture: a supervised \ac{CBM}-based Concept branch that delivers concept-level explainability, and a Visual branch that provides pixel-level anomaly localization. This combination pairs the semantic richness of concept-based explanations with the spatial precision of anomaly heatmaps. 

\item \textbf{Cross-dataset robustness evaluation.}
We validate our approach across three established industrial \ac{VAD} benchmarks, MVTec AD, VisA, and Real-IAD, demonstrating the generalizability and robustness of the proposed method across diverse industrial domains and defect types.

\end{enumerate}

The rest of the paper is organized as follows:
Sec.~\ref{sec:related_work} reviews relevant \ac{VAD} literature, while Sec.~\ref{sec:cbm} provides background on the \ac{CBM} architecture and training methodology.
Sec.~\ref{sec:methodology} presents our method, Concept-Aware \ac{VAD} (\ourmethod), detailing its adaptation of \acp{CBM} to the \ac{VAD} setting.
Sec.~\ref{sec:exp_setting} outlines the experimental setting with implementation details and considered supervision scenarios.
Sec.~\ref{sec:results} discusses the results, and Sec.~\ref{sec:conclusion} suggests directions for future work.
\section{Related Work}%
\label{sec:related_work}

Extensive research in Visual Anomaly Detection (VAD) has led to the development of numerous methods, which can be broadly categorized into four groups:
\\
\textbf{1. Feature-based Methods}: these approaches rely on features extracted from pre-trained neural networks. By leveraging the rich semantic and structural information encoded in these features, anomalies can be detected as deviations from normal patterns. Examples include PatchCore~\cite{patchcore}, STFPM~\cite{stfpm}, FastFlow~\cite{yu2021fastflow} and PaDiM~\cite{padim}.
\\
\textbf{2. Reconstruction-based Methods}: these methods assume models trained only on normal data cannot accurately reconstruct anomalous regions, making reconstruction errors indicative of anomalies. Examples include AnoGAN~\cite{anogan}, f-AnoGAN~\cite{fanogan}, and UniAD~\cite{you2022unified}.
Compared to feature-based methods, they require training generative models, which is computationally expensive. 
\\
\textbf{3. Synthetic Anomaly Methods}: these approaches augment datasets with synthetically generated anomalies to learn explicit anomalous patterns. Early methods, such as CutPaste~\cite{cutpaste}, DRAEM~\cite{zavrtanik2021draem}, and SuperSimpleNet~\cite{rolih2024supersimplenet}, relied on basic or unrealistic augmentations, while recent approaches leverage advanced generative models and \acp{VLM} to produce higher-fidelity anomalies, improving robustness and performance~\cite{sun2025unseen,hu2026mirage}. Notably, the generative model adopted by \ourmethod{} has been validated for realistic industrial anomaly synthesis~\cite{hu2026mirage}, supporting its suitability for this task.

A less explored setting is the semi-supervised one, with methods such as DevNet~\cite{pang2021explainable} and DRA~\cite{ding2022catching}, which instead leverage a small number of real anomalous examples alongside normal data, without relying on synthetic generation.

Although these methods provide visual explanations in the form of anomaly maps, they cannot generate semantic human-understandable explanations.
\\
\textbf{4. {VLM}-based anomaly detection.} Recent \acp{VLM}-based approaches combine anomaly detection with natural language description of anomalies~\cite{wu2024vadclip,gu2024anomalygpt,Huang_2024_CVPR,xu2025towards,zhu2024llms}.
While achieving strong zero-shot performance, which we do not target here, they require \ac{VLM} inference at the test time, introducing significant computational overhead that may limit industrial deployment. Moreover, zero-shot approaches rely on prior semantic knowledge rather than dataset statistics, potentially classifying semantically unusual but dataset-common patterns as anomalies. For instance, a dataset-wide discoloration may still be flagged as anomalous by a \ac{VLM}-based method, whereas \ourmethod{} and traditional methods would treat it as normal. In addition, their textual explanations are often post-hoc and may not faithfully reflect the factors driving the prediction, leading to hallucinated or inconsistent explanations~\cite{turpin2023language}. In contrast, \ourmethod{} enforces a semantic bottleneck between visual features and predictions, grounding decisions in explicit interpretable concepts and enabling interventions.


A different direction for improving interpretability is the use of concept-based explanations. Several works explore concept-based methods for detecting and explaining \ac{OOD} samples in computer vision~\cite{choi_concept-based_2023,sevyeri_transparent_2023,liu_component-aware_2023}. However, these methods typically define anomalies at the class level, e.g., treating images of cats as anomalous when the model is trained on dogs. This setting differs from realistic \ac{VAD} scenarios, where anomalies are usually subtle defects within otherwise normal objects. To the best of our knowledge, no prior work has investigated concept-based methods for \ac{VAD}. In this work, we address this gap by exploring the use of \acp{CBM} for \ac{VAD} and evaluating their effectiveness.

\section{Preliminaries: \aclp{CBM}}\label{sec:cbm}

A \ac{CBM}~\cite{koh2020concept} is an interpretable neural architecture that makes predictions via an intermediate set of human-interpretable concepts. A key advantage is that \acp{CBM}, unlike large \ac{VLM}, require very limited memory and enable fast inference while still providing human-understandable explanations, making them suitable for resource-constrained or real-time settings. 
Formally, let $\mathbf{x} \in \mathcal{X}$ be the input, $\mathbf{c}\in \mathcal{C} $ a vector of $k$ concepts, and $y \in \mathcal{Y}$ the target label. A \ac{CBM} consists of two functions: a concept extractor $g: \mathcal{X} \rightarrow \mathcal{C}$, which maps the input $\mathbf{x}$ to a predicted concept vector $\mathbf{\hat{c}}=g(\mathbf{x})$, and a label predictor $f: \mathcal{C} \rightarrow \mathcal{Y}$, which maps the predicted concepts $\mathbf{\hat{c}}$ to the final output $\hat{y} = f(\mathbf{\hat{c}})$.
The \ac{CBM} is then the composition $\hat{y} = f(g(\mathbf{x})) $.
\\
The concept layer acts as a bottleneck, forcing predictions $\hat{y}$ to depend on $\mathbf{x}$ entirely through the concept predictions. This improves the interpretability of the downstream output and enables test-time intervention, as discussed below.

Standard \acp{CBM} require supervised learning, i.e., require access to ground-truth triplets $\{\mathbf{x}^i, \mathbf{c}^i, y^i\}_{i=1}^m$ during training.
In our setting, the input $\mathbf{x}$ is a raw image, $y\in  \{0,1\}$  indicates whether an anomaly is present, and true concepts are binary, i.e., $\mathcal{C}={\{0,1\}}^{k}$.

\subsection{\ac{CBM} Training}\label{subsec:cbmtraining}

A \ac{CBM} can be trained using different strategies. In this work, we adopt joint end-to-end training, which consistently outperformed alternative paradigms (independent and sequential training). Details and ablation studies for the alternative paradigms are provided in the Supplementary Material.

In \textbf{Joint training}, the concept extractor $g$ and label predictor $f$ are optimized jointly by minimizing
\begin{equation}
    \hat{f}, \hat{g} = \arg \min_{f,g} \sum_i \mathcal{L}_Y(f(g(\mathbf{x}^i)), y^i) + \lambda \sum_{i,j} \mathcal{L}_{C_j}(g_j(\mathbf{x}^i), c_j^i),
\end{equation}
where $\mathcal{L}_Y$ is the label prediction loss, $\mathcal{L}_{C_j}$ is the loss for the $j$-th concept, and $\lambda$ balances the two objectives.

Note that the predicted concepts $g_j(\mathbf{x}^i)$ are continuous logits rather than binary variables as in the ground-truth concept vectors. The classifier $f$ operates directly on these logits.

\subsection{\ac{CBM} Test-time Interventions}
A key property of \acp{CBM} is their support for test-time interventions. When a user or domain expert identifies incorrectly predicted concepts, they can manually correct them by setting $\hat{c}_j=c_j$. The resulting correct concept vector $\mathbf{\tilde{c}}$ is fed into the label predictor $f$, obtaining an updated (and ideally more accurate) output $\tilde{y} = f(\mathbf{\tilde{c}})$.
In jointly trained models, where $f$ operates on concept logits, interventions replace the corresponding logits with the 5th or 95th percentile of the training logit distribution to avoid extreme values (see Supplementary Material for more details).
This mechanism allows experts to improve predictions without retraining the model, a capability enabled by the \acp{CBM} bottleneck structure.

\begin{figure}[t]
\centering
\includegraphics[width=\linewidth]{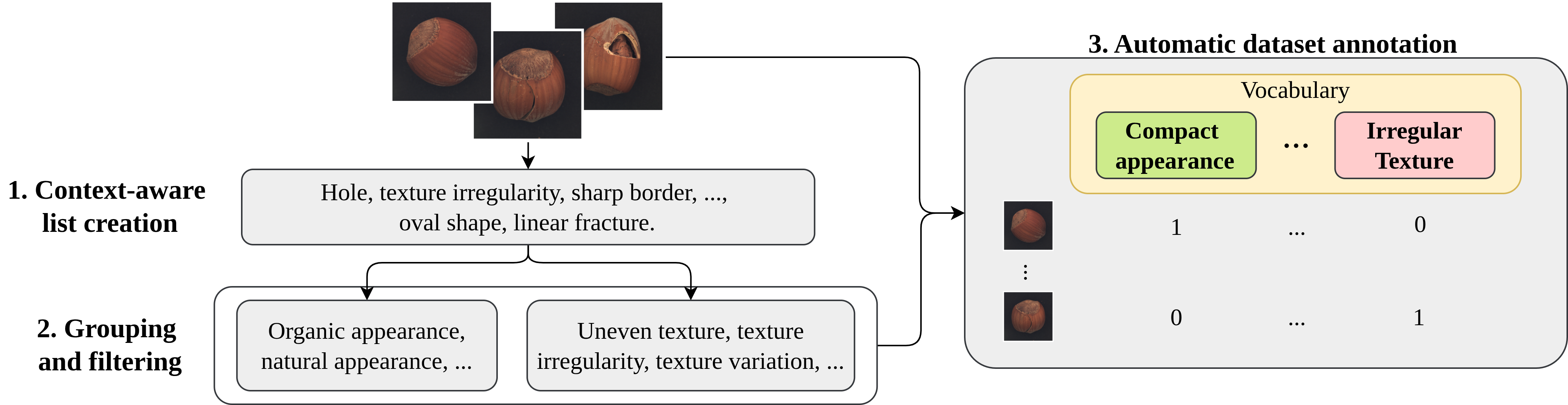}
\caption{Pipeline for creating the Concept Dataset through concept annotation of a \ac{VLM}.}%
\label{fig:concept_dataset_annotation}
\end{figure}

\section{Methodology}\label{sec:methodology}

\subsection{Concept Dataset Pipeline}

A key challenge of \acp{CBM} is the lack of concept-annotated datasets, which are unavailable in many domains, including \ac{VAD}. Prior works addressed this using automated pipelines based on \acp{VLM}, such as Label-free \acp{CBM}~\cite{oikarinen2023label}, LaBo~\cite{yang2023language}, and VLG-\acp{CBM}~\cite{srivastava2024vlg}.
However, our setting differs from prior work, which focused on multiclass classification of natural images. In \ac{VAD}, concepts are domain-specialized and fine-grained. Moreover, industrial anomalies are highly dataset- and category-dependent and are underrepresented in \ac{VLM} training data, which can lead to generic or hallucinated concepts. 
We therefore devise a dedicated anomaly-oriented concept extraction pipeline and prompt (see details in the Supplementary Material), and we validate it against a manually annotated ground truth. Our approach draws inspiration from existing concept-extraction methods in other domains, with minimal adaptations for \ac{VAD}.

Our procedure, shown in Fig.~\ref{fig:concept_dataset_annotation}, consists of three steps:

\smallskip
\noindent \textbf{1. Context-aware list creation.}
Unlike existing approaches, which generate concepts without domain context, we introduce a context-aware paradigm tailored to each object category.
We sample a representative subset $X = \{x_i \in D \mid i=1,\dots,N\}$ (5\% of the dataset) ensuring coverage of all defect types.
Let $\mathcal{M}$ be the \ac{VLM} and $pt$ a textual prompt that injects contextual information (object category, defect type) and employs a Chain-of-Thought strategy \cite{Wei2022ChainOT} (see Supplementary Material for details on the prompt design). The model first generates a natural language image description and then extracts up to five semantic concepts from it. By constraining extraction to a visual description, we prevent the model from focusing on non-visual attributes and limit hallucinations. The resulting concept set is:
\begin{equation}
\mathcal{V} = \{v^i_j \mid i = 1,\dots,N;\ j = 1,\dots,|v^i|\}, \qquad v^i = \mathcal{M}(x^i, pt).
\end{equation}

\noindent \textbf{2. Grouping and filtering.}
The raw concepts are refined in two substeps:
(i)~\textbf{Concept grouping}: the full list is fed back to the \ac{VLM} with few-shot examples \cite{brown2020language} to merge morphologically and semantically related attributes, reducing the set from a few hundred to fewer than fifty concepts.
(ii)~\textbf{Concept filtering}: following \cite{oikarinen2023label}, we encode concepts with the CLIP ViT-B/32 text encoder $E_T$ \cite{radford2021learning} and remove near-duplicates by discarding one concept from each pair exceeding a cosine similarity of 0.9:
\begin{equation}
\cos(E_T(v_i), E_T(v_j)) = \frac{E_T(v_i)\cdot E_T(v_j)}{\|E_T(v_i)\|\ \|E_T(v_j)\|}
\end{equation}

\noindent \textbf{3. Automatic dataset annotation.}
Each image is annotated with binary concept labels. A \ac{VLM} is queried with a category- and defect-aware prompt to inspect each image against the final concept list and return a boolean assignment per concept (see Supplementary Material). The quality of the automatic annotation process is tested against a manually annotated subset of 400 images across all categories, achieving 91.3\% accuracy (see Supplementary Material for further details).

Fig.~\ref{fig:concept_distribution_example} shows the resulting vocabulary and its distribution for the Hazelnut category in MVTec, divided by defect types. Concepts exhibit meaningful differences between the normal and anomalous class, confirming their informativeness for the downstream task.
Anomalous-class concepts include both general indicators (\textit{uneven tone, surface discontinuity}) and fine-grained descriptors (\textit{visible ink, dark interior, hole}), with the latter exhibiting distinct distributions across different defect types and therefore enabling more precise and discriminative characterization of each specific defect.
Notably, concepts are not exclusive to normal or abnormal classes; many appear in both with different frequencies. Thus, there is no one-to-one mapping between individual concepts and the final label, and classification relies on their joint interactions rather than individual concepts.

\begin{figure}[thbp]
\centering
\includegraphics[width=0.9\linewidth]{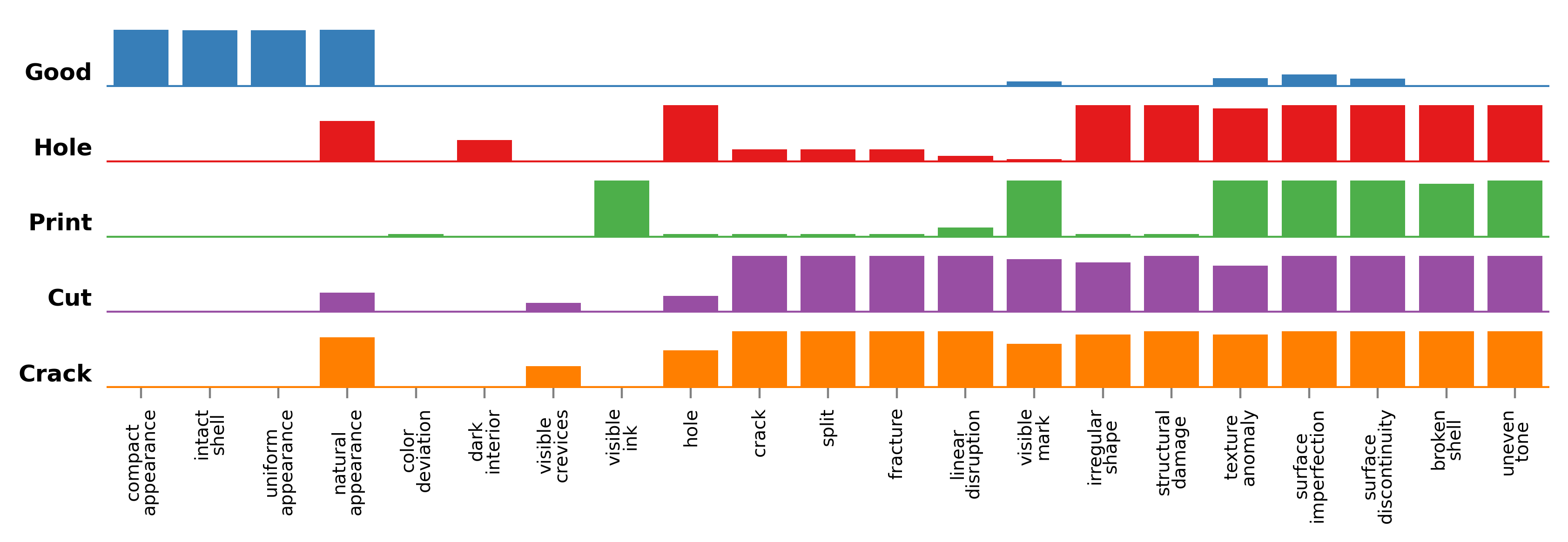}
\caption{Example of concept vocabulary and distribution across defect types for the MVTec hazelnut category.}%
\label{fig:concept_distribution_example}
\end{figure}

\subsection{\ourmethod{} architecture}
Modern \ac{VAD} approaches provide pixel-level anomaly localization, whereas standard \acp{CBM} cannot. We bridge this gap by augmenting \acp{CBM} with a spatial localization branch, enabling them to jointly produce concept-based explanations and fine-grained anomaly heatmaps.
The proposed architecture, illustrated in Fig.~\ref{fig:architecture}, consists of two branches: (i) a \emph{Concept branch} $\phi_{\text{cbm}}$, which produces a concept prediction and an anomaly score, and (ii) a \emph{Visual branch} $\phi_{\text{vis}}$, trained as a student to match the Concept-branch features on normal data and producing spatial anomaly localization together with an additional anomaly score. The two anomaly scores are merged as in Eq.~\ref{eq:final_score}. Sharing the feature extractor across branches encourages consistency between semantic and spatial explanations.

\subsubsection{Training}
Training proceeds in two sequential phases. First, $\phi_{\mathrm{cbm}}$ is trained end-to-end on concept prediction and anomaly detection objectives. Then, its weights are frozen and $\phi_{\mathrm{vis}}$ is trained as a student to match the intermediate features of $\phi_{\mathrm{cbm}}$ on normal samples only, following the student--teacher \ac{VAD} paradigm~\cite{stfpm,zhang2023destseg,rd4ad,barusco2025paste}.

\smallskip
\noindent\textbf{Concept Branch.}
We define per-concept and anomaly detection losses:

\begin{equation}
\mathcal{L}_{C_j} = \mathcal{L}_{\text{BCE}}(c_j,\hat{c}_j,\alpha),
\qquad
\mathcal{L}_{Y} = \mathcal{L}_{\text{BCE}}(y,\hat{y},\alpha).
\end{equation}

\noindent where $\mathcal{L}_{\text{BCE}}(z,\hat{z};\alpha) =-\frac{1}{n} \sum_{i=1}^n (\alpha z^i \log \hat{z}^i + (1-z^i) \log (1-\hat{z}^i))$.
\\
\noindent The weighting factor $\alpha \in \mathbb{R}$ compensates for class imbalance in both concept and anomaly predictions, and is computed as the imbalance ratio $\alpha = (n - \sum_{i=1}^n z^i)/(\sum_{i=1}^n z^i)$.
The two losses are combined as:
\begin{equation}
    \mathcal{L}_{cbm} = \frac{1}{1+\lambda k}\left(\mathcal{L}_Y + \lambda\sum_{j=1}^{k}\mathcal{L}_{C_j}\right)
\end{equation}
where $k$ denotes the number of concepts and $\lambda \in \mathbb{R}$ controls the trade-off between the anomaly detection and concept prediction objectives.

\smallskip
\noindent\textbf{Visual Branch.}
Once $\phi_{\mathrm{cbm}}$ is frozen, $\phi_{\mathrm{vis}}$ is optimized to match its intermediate feature representations on normal samples only:

\begin{equation}
\mathcal{L}_{\text{vis}} = \frac{1}{|\mathcal{X}_n|} \sum_{\mathbf{x} \in \mathcal{X}_n}
\frac{1}{F} \sum_{f=1}^{F}
\left\| \phi_{\text{cbm}}^{(f)}(\mathbf{x}) - \phi_{\text{vis}}^{(f)}(\mathbf{x}) \right\|_2^2,
\end{equation}
where $\mathcal{X}_n$ denotes the set of normal training samples, $F$ is the fixed number of intermediate feature layers~\cite{stfpm,barusco2025paste}, and $\phi_{\text{cbm}}^{(f)}, \phi_{\text{vis}}^{(f)}$ indicate the $f$-th layer of the respective network.

\subsubsection{Inference}

Given a test image $\mathbf{x}$, the model produces three outputs: a concept-based explanation, a spatial heatmap and an anomaly score.

\smallskip
\noindent \textbf{Concept-based Explanation.} Given the predicted concept activations 
$\hat{\mathbf{c}} = (\hat{c}_1, \dots, \hat{c}_k)$ and the learned weights 
$\mathbf{w} = (w_1, \dots, w_k)$ of the final linear layer, the contribution of 
each concept $c_j$ to the anomaly score is proportional to:
\begin{equation}
    \delta_j(\mathbf{x}) = w_j \cdot \hat{c}_j(\mathbf{x}).
\end{equation}
The explanation is then presented as the ranked list of concepts sorted by 
$|\delta_j(\mathbf{x})|$, highlighting those that most influenced the final 
prediction.

Compared to classic VAD approaches, \ourmethod{} has two key advantages. First, the model is inherently interpretable, as ranked concept contributions provide a direct, human-readable justification for the anomaly prediction. Second, the most relevant concept activations $\hat{\mathbf{c}}$ can be exposed to human experts for validation, enabling human-in-the-loop intervention by correcting mispredicted concepts and directly influencing the final decision without retraining~\cite{chauhan2023interactive}.

\smallskip

\noindent \textbf{Anomaly Heatmap.}
At inference, both networks process the input image $\mathbf{x}$, and an anomaly heatmap is
computed from the discrepancy between their feature maps, as anomalous regions
lie outside the student training distribution:
\begin{equation}
H_{\text{vis}}(\mathbf{x}) = \frac{1}{F}\sum_{f=1}^{F} \mathcal{U}\!\left(D^{(f)}(\mathbf{x})\right), \qquad
D^{(f)}_{\text{h,w}}(\mathbf{x}) = \left\|\phi^{(f)}_{\text{cbm,h,w}}(\mathbf{x})-\phi^{(f)}_{\text{vis,h,w}}(\mathbf{x})\right\|_2^2,
\end{equation}
where $\mathcal{U}(\cdot)$ denotes bilinear upsampling to the original resolution
$H \times W$ and the per-layer squared difference is averaged across feature channels.
This approach pairs the textual explanations produced by the \ac{CBM} branch with 
a spatial visualization of anomalous regions. A schematic illustration is shown in 
Fig.~\ref{fig:architecture}.

\smallskip
\noindent \textbf{Anomaly Score.}
To align the two branches onto a common scale, we calibrate the Visual branch using a small held-out set $\mathcal{X}_\mathrm{cal}$. The per-image visual score is obtained by max-pooling the heatmap over spatial dimensions, $s_{\mathrm{vis}}(\mathbf{x}) = \max_{h,w}\, \mathbf{H}_{\mathrm{vis}}^{(h,w)}(\mathbf{x})$, and the threshold $\tau^*$ maximising F1 on $\mathcal{X}_\mathrm{cal}$ yields the binarized prediction $\hat{y}_\mathrm{vis}(\mathbf{x}) = \mathbf{1}[s_\mathrm{vis}(\mathbf{x}) \geq \tau^*] \in \{0,1\}$.
The Concept branch outputs a sigmoid anomaly probability $s_\mathrm{cbm}(\mathbf{x}) = \sigma(\phi_\mathrm{cbm}(\mathbf{x})) \in [0,1]$ conditioned on $\hat{\mathbf{c}}(\mathbf{x})$.
The final score combines both signals:
\begin{equation}\label{eq:final_score}
    s(\mathbf{x}) = \max\!\bigl(\hat{y}_{\mathrm{vis}}(\mathbf{x}),\; s_{\mathrm{cbm}}(\mathbf{x})\bigr).
\end{equation}
An anomaly detected by either branch is sufficient to flag the image, making detection more resilient to failures of a single branch.
Notably, as shown in the right subtable of Table~\ref{tab:united_results}, the Concept branch drives the final prediction in approximately 88--91\% of cases across supervision levels, meaning the Visual branch acts primarily as a fallback for out-of-distribution samples not captured by the concept layer. As a result, the vast majority of detections are driven by concept-level explanations. An ablation regarding alternative fusion strategies is available in the Supplementary Material, displaying that, while achieving a higher AUROC, they consistently reduce agreement with the concept branch.

\subsection{Synthetic Anomaly Generation}

In line with prior works~\cite{sun2025unseen,hu2026mirage}, we adopt a generative approach for realistic synthetic anomaly generation. By incorporating these synthetic samples, \ourmethod{} can operate across all supervision regimes, including semi-supervised and unsupervised settings where no real anomalous images are available.
\\
In detail, synthetic anomalous images are created by modifying normal images using a text-to-image and image-editing system.
Let $\mathcal{G}_{\text{edit}}(\mathbf{x}, p)$ denote the editing function, which takes
$\mathbf{x} \in \mathbb{R}^{H \times W \times 3}$ as input image and $p$ that represents a textual prompt.
Starting from a normal image $\mathbf{x}_n$, the generative procedure creates an anomalous image $ \mathbf{x}_a = \mathcal{G}_{\text{edit}}(\mathbf{x}_n, p_a)$
where $p_a$ includes the desired defect type and ensures consistency with the original image.
The scope of this operation is to introduce anomalies while preserving the rest of the image unchanged.

Prompts are carefully designed to control the generation of anomalies (see Supplementary Material for further details). 
Each prompt specifies:
(i) Defect type: the type of anomaly to introduce (e.g., scratch, crack, dent).
(ii) Object type: the object type or surface being modified (e.g., metal plate, plastic part).
(iii) Pose and view details: information that ensures the edited image maintains the same camera angle, orientation, and lighting as the original.

\section{Experimental Setting}\label{sec:exp_setting}

\subsection{Scenarios and baselines}
In this work, we evaluate \ourmethod{} under several scenarios that differ in the 
type and amount of supervision provided during training. We adopt the same definitions as \cite{xie2024iad}. 
\\
\textbf{Supervised}: training on both normal and anomalous images from the real-world dataset. We compare against supervised \ac{AD} based on UNet \cite{ronneberger2015u}, due to the absence of specific fully-supervised \ac{VAD} methods that train on the same datasets and training conditions.
\\
\textbf{Semi-Supervised}: training on all available normal images and synthetically generated anomalies, together with a limited number of real anomalous samples. Specifically, we adopt a one-shot setting per defect type, resulting in an average of 5 real anomalous images per category. For comparison, we evaluate against the semi-supervised approaches DevNet~\cite{pang2021explainable} and DRA~\cite{ding2022catching} under the same setting, using 5 anomalous images per category.
\\
\textbf{Unsupervised}: training uses only normal images from the real-world dataset, together with synthetically generated samples. 
Following established prior works \cite{xie2024iad,liu2023simplenet,cutpaste}, this setting is still considered unsupervised since no real anomalous samples are used during training. In this scenario, concepts are extracted exclusively from synthetically generated anomalous images, so the model never accesses real defective samples during concept extraction; moreover, calibration of the F1-maximizing threshold for score fusion uses only synthetic anomalies and normal images, while inference is performed on real anomalous images, keeping calibration and test data fully disjoint. We compare against PatchCore \cite{patchcore}, STFPM \cite{stfpm}, Tiny-Dinomaly \cite{barusco2026continual}, AnomalyGPT \cite{gu2024anomalygpt} and the zero-shot method Anomaly-OV \cite{xu2025towards}. Note that Anomaly-OV operates in a fully zero-shot manner, requiring no training images whatsoever, whereas AnomalyGPT, despite being VLM-based, requires normal reference images at inference time and therefore falls under the unsupervised scenario.

\subsection{Implementation Details}

\textbf{Models.} 
In the Concept Dataset Annotation Pipeline, we mainly leverage Gemma3 \cite{team2025gemma} as a \ac{VLM}, except in the Concept Filtering phase, where we use the CLIP ViT-B/32 text encoder \cite{radford2021learning}.
For the text-to-image and image-editing system in the synthetic anomaly generation pipeline, we employ Gemini 2.5 Flash Image (Nano Banana). The \ac{CBM} uses MobileNet-v2 as a feature extractor \cite{sandler2018mobilenetv2}. 
\\
PatchCore and STFPM are evaluated using the same MobileNet backbone as our approach, similarly to the UNet, which was built using MobileNet-v2 as an encoder to ensure comparability. Tiny-Dinomaly adopts a lightweight DeiT backbone, while AnomalyGPT uses Vicuna-7B with ImageBind-Huge as image encoder, and Anomaly-OV uses LLaVA\--One\-Vision 0.5B, as in the original papers.
\\
\textbf{Datasets.} Our main experimental results are obtained using the \ourdataset\ dataset, and further results on industrial benchmarks, including VisA and Real-IAD, confirm cross-dataset robustness.
\\
\textbf{Hyperparameter Optimization.} 
To select the best hyperparameter configuration, we employ Bayesian optimization, implemented through the Optuna library~\cite{akiba2019optuna}, using the Tree-structured Parzen Estimator (TPE) algorithm~\cite{watanabe2023tree} (details in Supplementary Material).
\\
\textbf{Training Details.}
\ourmethod{} results are averaged over three runs with different random seeds. Inference uses a held-out set composed of 20\% of the real-world dataset.
Additional details about training and data augmentation in Supplementary Material. The full implementation is available in the open-source library MoViAD \cite{barusco2025moviad}.

\subsection{Metrics}\label{subsec:metrics}

VAD algorithms can be assessed using multiple criteria, corresponding to different levels of analysis:
\\
\textbf{Image-level:} To evaluate detection at the image level (each image is a different sample), we report the AUROC (Area under the Receiver Operating Characteristic curve) score, referred to as I-AUC.
\\
\textbf{Pixel-level:} 
To assess performance at the pixel level, we consider the commonly used AUROC score, named P-AUC, computed by considering each pixel of each image as an independent sample.
\\
\textbf{Concept-level:} For \ourmethod{}, we evaluate the accuracy of concept predictions using the average AUROC score, referred to as C-AUC.
Since concepts are binary, we compute the AUROC score for each concept separately and then combine them by averaging over all concepts. 
Each term is weighted by $w_j$, i.e., the number of true instances for the $j-$th concept $c_j$, where $k$ is the total number of concepts:  
\begin{equation}
\label{eq:cauc}
\text{C-AUC} := \frac{\sum_{j=1}^{k} w_j \cdot \text{AUROC}(c_j)}{\sum_{j=1}^{k} w_j},
\end{equation}

\begin{table}[thbp]
\centering
\caption{Comparison of \ourmethod{} and other VAD methods across unsupervised, semi-supervised, and supervised scenarios on the MVTec AD dataset. \xmark{} indicates that the method cannot produce the corresponding output (pixel localization or concept-level prediction).
}
\label{tab:comparison}
\resizebox{0.98\columnwidth}{!}{%
\begin{tabular}{
l
>{\centering\arraybackslash}c
>{\centering\arraybackslash}c
>{\centering\arraybackslash}c
>{\centering\arraybackslash}c
>{\centering\arraybackslash}c
>{\centering\arraybackslash}c
>{\centering\arraybackslash}c
>{\centering\arraybackslash}c
>{\centering\arraybackslash}c
>{\centering\arraybackslash}c
>{\centering\arraybackslash}c
}
\toprule
\multicolumn{1}{c|}{\textbf{Metric}} &
\multicolumn{1}{c|}{\textbf{Zero-Shot}} &
\multicolumn{5}{c|}{\textbf{Unsupervised (Zero-Anomaly-Shot)}} &
\multicolumn{3}{c|}{\textbf{Semi-Supervised}} &
\multicolumn{2}{c}{\textbf{Supervised}} \\
\multicolumn{1}{c|}{\textbf{}} &
\multicolumn{1}{c|}{\textbf{Anomaly-OV}} &
\textbf{CONVAD} &
\textbf{PatchCore} &
\textbf{STFPM} &
\textbf{Tiny-Dinomaly} &
\multicolumn{1}{c|}{\textbf{AnomalyGPT}} &
\textbf{CONVAD} &
\textbf{DevNet} &
\multicolumn{1}{c|}{\textbf{DRA}} &
\textbf{CONVAD} &
\textbf{UNet} \\
\midrule
\rowcolor{white}
\multicolumn{1}{l|}{I-AUC} &
\multicolumn{1}{c|}{0.94} &
0.81 &
0.86 &
0.88 &
0.91 &
\multicolumn{1}{c|}{\textbf{0.97}} &
\textbf{0.87} &
0.80 &
\multicolumn{1}{c|}{0.76} &
\textbf{0.96} &
{0.65} \\
\multicolumn{1}{l|}{P-AUC} &
\multicolumn{1}{c|}{\xmark} &
{0.91} &
{\textbf{0.96}} &
{0.88} &
{\textbf{0.96}} &
\multicolumn{1}{c|}{{0.93}} &
{\textbf{0.92}} &
{\textbf{0.92}} &
\multicolumn{1}{c|}{\xmark{}} &
\textbf{0.95} &
{0.77} \\
\rowcolor{white}
\multicolumn{1}{l|}{C-AUC} &
\multicolumn{1}{c|}{\xmark{}} &
\textbf{0.73} &
\xmark{} &
\xmark{} &
\xmark{} &
\multicolumn{1}{c|}{\xmark{}} &
\textbf{0.72} &
\xmark{} &
\multicolumn{1}{c|}{\xmark{}} &
\textbf{0.86} &
\xmark{} \\
\multicolumn{1}{l|}{\makecell[l]{Memory \\ {[MB] $\downarrow$}}} &
\multicolumn{1}{c|}{1{,}768} &
{17.6} &
38.4 &
\textbf{11.1} &
36.5 &
\multicolumn{1}{c|}{30{,}520} &
{\textbf{17.6}} &
44.7 &
\multicolumn{1}{c|}{44.7} &
{17.6} &
{\textbf{10.3}} \\
\bottomrule
\end{tabular}%
}
\end{table}

\vspace{-20pt}
\section{Results}\label{sec:results}

Sec.~\ref{subsec:cbm_scenarios} compares \ourmethod{} with state-of-the-art VAD approaches across supervision regimes.
Sec.~\ref{subsec:user_study} reports a user study assessing perceived explanation quality.
Sec.~\ref{subsec:intervention} evaluates concept-level interventions as a mechanism for human-machine collaboration.
Finally, Sec.~\ref{subsec:ablation} ablates the contribution of the Concept and Visual branches and analyzes the pixel-level performance of \ourmethod{}.

\subsection{CBM Scenarios}\label{subsec:cbm_scenarios}
\vspace{-1mm}
A comparison of \ourmethod{} against state-of-the-art VAD approaches on MVTec AD is presented in Table~\ref{tab:comparison} across supervised, semi-supervised, and unsupervised scenarios; additional results on VisA and Real-IAD datasets are reported in Table~\ref{tab:united_results} (Left).

\noindent\textbf{Supervised.}
When compared with the traditional UNet baseline in the supervised setting, \ourmethod{} achieves improved performance while additionally providing concept-based explanations that make predictions interpretable to domain experts.
Although the supervised setting has access to all available anomalous samples, these remain scarce, thereby limiting the performance of UNet and highlighting the data efficiency of our method. Further results in the supervised setting across the three considered benchmarks (MVTec AD, VisA, Real-IAD) confirm the effectiveness of \ourmethod{}.
\\
\ourmethod{} and UNet have a comparable memory footprint (17.6\,MB vs.\ 10.3\,MB); however, UNet achieves substantially lower detection performance (I-AUC: 0.65 vs.\ 0.96, P-AUC: 0.77 vs.\ 0.95) despite requiring full supervision, suggesting that its limited generalization makes it an impractical choice for real-world industrial inspection.
\\
\noindent\textbf{Semi-supervised.}
\ourmethod{} outperforms DevNet and DRA, two established semi-supervised VAD baselines. This suggests that the concept-based inductive bias embedded in \ourmethod{} acts as an effective regularizer, improving resilience to scarce anomalous training data and generalization under limited supervision. A key enabler is the synthetic anomaly generation procedure, which produces samples that closely align with the real abnormal domain (Fig.~\ref{fig:tsne_metalnut}), providing the model with more representative concept supervision despite limited real anomalous data. See Sec.~\ref{subsec:sim2real} for details.
\\
\noindent\textbf{Unsupervised.}
\ourmethod{} reaches an I-AUC of 0.81, trailing the similarly-sized STFPM (0.88) and Tiny-Dinomaly (0.91), the latter requiring twice the memory (36.5\,MB vs.\ 17.6\,MB). We consider this an acceptable cost, as none of these baselines exposes concept-level explanations and the gap is largely recovered through interventions (Sec.~\ref{subsec:intervention}).
Methods with higher detection scores come with significant trade-offs: AnomalyGPT achieves the best I-AUC (0.97) but requires over 30\,GB of memory (nearly $1{,}750\times$ more than \ourmethod{}) and ranks as the worst method in our user study on explanation quality (Sec.~\ref{subsec:user_study}), with 44.2\% of its explanations rated as unhelpful.
Similarly, Anomaly-OV attains strong detection performance (I-AUC: 0.94) but produces no pixel-level heatmaps, which are a fundamental output in industrial VAD applications. 
In contrast, \ourmethod{} delivers all three outputs: concept explanations, pixel heatmaps, and anomaly scores with a compact footprint, and benefits from human-in-the-loop interventions that can boost detection performance at inference time (Sec.~\ref{subsec:intervention}).

\begin{table}[thbp]
    \centering
    \caption{(Left) Average results on MVTec AD, VisA, and Real-IAD (Supervised) for the Concept branch vs.\ PatchCore (WideResNet-50, although unsupervised, is a strong baseline) and AnomalyGPT (4-normal-shot). (Right) Ablation of \ourmethod{} on MVTec AD across supervision levels; I-AUC reported for the Concept branch, Visual branch, and their fusion. Concept Rate is the fraction of samples where the Concept branch drives the final prediction.}%
    \label{tab:united_results}
    \begin{minipage}[t]{0.46\linewidth}
        \centering
        \resizebox{\linewidth}{!}{%
        \begin{tabular}{l c c c c}
            \midrule
                             & \multicolumn{2}{c}{\ourmethod{}} & PatchCore & AnomalyGPT \\
            \textbf{Dataset} & \textbf{C-AUC} & \textbf{I-AUC} & \textbf{I-AUC} & \textbf{I-AUC} \\
            \midrule
            MVTec AD         & 0.86           & 0.97           & \textbf{0.99} & 0.96          \\
            VisA             & 0.76           & \textbf{0.94}  & 0.93          & 0.91          \\
            Real-IAD         & 0.88           & \textbf{0.94}  & 0.90          & ---            \\
            \midrule
        \end{tabular}
        }
    \end{minipage}%
    \hspace{8pt}%
    \begin{minipage}[t]{0.46\linewidth}
        \centering
        \resizebox{\linewidth}{!}{%
        \renewcommand{\arraystretch}{0.84}%
        \setlength{\tabcolsep}{6pt}%
        \begin{tabular}{l c c c c}
            \midrule
            \textbf{Sup.\ Level} & \multicolumn{3}{c}{\textbf{I-AUC}} & \textbf{Concept} \\
            & \textbf{Concept} & \textbf{Visual} & \textbf{Fusion} & \textbf{Rate (\%)} \\
            \midrule
            Unsuper.    & 0.77  & 0.90  & 0.81  & 88.3  \\
            Semi-Sup.   & 0.85  & 0.91  &  0.87 & 90.1  \\
            Supervised  & 0.97  & 0.96  & 0.96  & 91.3  \\
            \midrule
        \end{tabular}
        }
    \end{minipage}
\end{table}

\subsection{User Study}\label{subsec:user_study}
\begin{table}[thbp]
\centering
\caption{User study results (582 responses, 32 participants). \textbf{Left}: perceived helpfulness of explanations. \textbf{Right}: explanation quality signals; \textit{effective}: the explanation enabled the user to verify whether the prediction was correct; \textit{misleading}: it convinced the user to trust a wrong prediction (counting defined in Eq.~\ref{eq:quality_signals}).}
\label{tab:user_study}
\small
\setlength{\tabcolsep}{4pt}
\begin{minipage}[t]{0.48\linewidth}
\centering
\begin{tabular}{lccc}
\toprule
Method & Helpful & Partial & Unhelpful $\downarrow$ \\
\midrule
AnomalyGPT~\cite{gu2024anomalygpt} & 33.2\% & 22.6\% & 44.2\% \\
Anomaly-OV~\cite{xu2025towards}    & \textbf{59.2\%} & 22.5\% & 18.3\% \\
\textbf{\ourmethod{}}                      & 55.6\% & 28.6\% & \textbf{15.8\%} \\
\bottomrule
\end{tabular}
\end{minipage}
\hfill
\begin{minipage}[t]{0.48\linewidth}
\centering
\begin{tabular}{lccc}
\toprule
Signal & AGPT & A-OV & \ourmethod{} \\
\midrule
Effective  & 33.2\% & 60.7\% & \textbf{73.5\%} \\
Misled $\downarrow$ & 18.4\% & 11.7\% & \textbf{3.1\%}  \\
Useless $\downarrow$   & 48.4\% & 27.6\% & \textbf{23.5\%} \\
\bottomrule
\end{tabular}
\end{minipage}
\end{table}

We conducted an online study with 32 participants to assess perceived explanation quality on MVTec and VisA images (See Supplementary Material for more details). Each participant was shown a product image with the model prediction, its explanation and, when available, the anomaly heatmap, then asked (Q1) if the prediction appeared correct and (Q2) if the explanation was helpful. Responses were collected for \ourmethod{}, AnomalyGPT~\cite{gu2024anomalygpt}, and Anomaly-OV~\cite{xu2025towards}, balanced across methods and categories (582 responses total).
\\
We combine both answers into a single quality signal. For a response, let $q_1\in\{\text{yes},\text{no},\text{unsure}\}$ encode the user's verdict on the prediction (Q1), $q_2\in\{\text{yes},\text{partial},\text{no}\}$ the perceived helpfulness (Q2), and $m\in\{0,1\}$ whether the prediction actually matched the ground truth (logged server-side, hidden from the user). An explanation is \emph{effective} when it is helpful and the user's verdict agrees with the truth, and \emph{misleading} when it is helpful but steers the user against the truth. Over $N$ responses,
\begin{equation}\label{eq:quality_signals}
\begin{aligned}
\mathrm{Effective}&=\tfrac{1}{N}\big|\{\,q_1{\neq}\text{unsure}\,\wedge\,q_2{\neq}\text{no}\,\wedge\,(q_1{=}\text{yes}\Leftrightarrow m{=}1)\,\}\big|\\
\mathrm{Misled}&=\tfrac{1}{N}\big|\{\,q_1{\neq}\text{unsure}\,\wedge\,q_2{\neq}\text{no}\,\wedge\,(q_1{=}\text{yes}\Leftrightarrow m{=}0)\,\}\big|,
\end{aligned}
\end{equation}
and the remainder is \emph{useless} ($q_2{=}\text{no}$ or $q_1{=}\text{unsure}$), so $\mathrm{Useless}=1-\mathrm{Effective}-\mathrm{Misled}$.
\\
In safety-critical industrial inspection, a useful explanation must enable the operator to independently verify whether the prediction is correct, catching both false alarms and missed defects before they propagate downstream.
\\
As reported in Table~\ref{tab:user_study}, \ourmethod{} rates as helpful or partially helpful in 84.2\% of cases, on par with Anomaly-OV (81.7\%) despite a slightly lower outright helpfulness rate (55.6\% vs.\ 59.2\%). More importantly, \ourmethod{} achieves the lowest unhelpful rate (15.8\% vs.\ 18.3\% and 44.2\%) and the fewest misleading explanations (3.1\% vs.\ 11.7\% and 18.4\%), indicating that Anomaly-OV's marginal helpfulness advantage comes at the cost of substantially greater risk of misleading operators.
The intrinsic coupling between concept scores and the anomaly prediction, as opposed to post-hoc verbalization in VLM-based methods, reduces noise in the explanatory signal and prevents errors from being masked by plausible-sounding text.

\subsection{Intervention}\label{subsec:intervention}
\begin{figure}[thbp]
    \centering
    \includegraphics[width=0.85\linewidth]{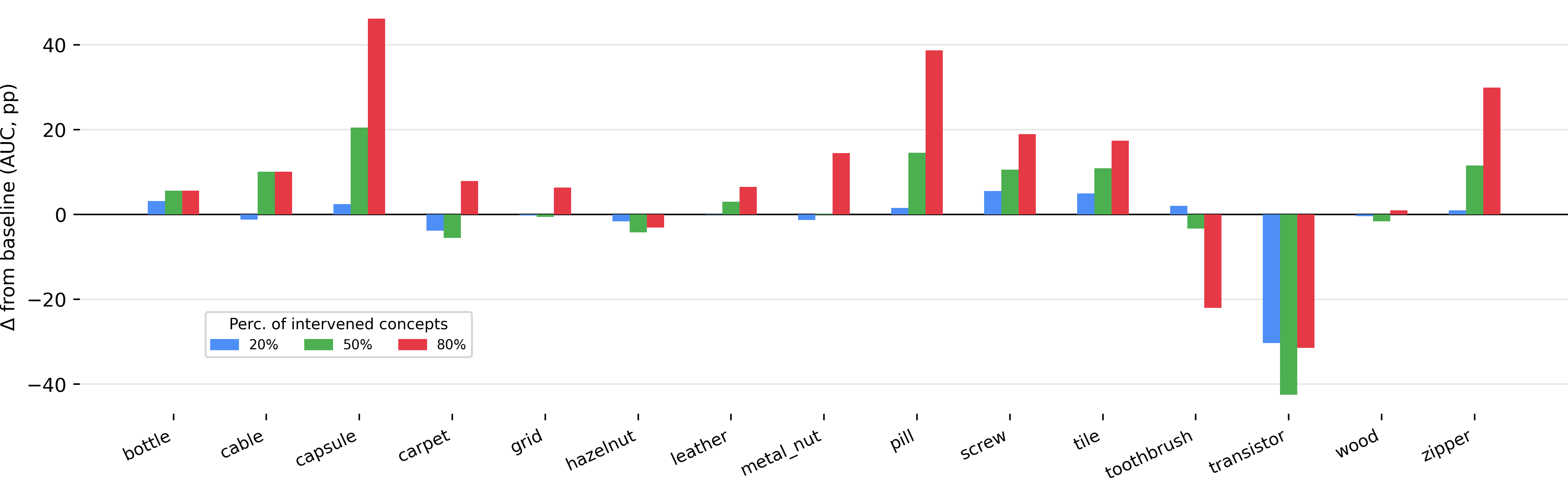}
    \caption{Performance gain in MVTec categories for increasing number of intervened concepts. Unsupervised setting.}%
    \label{fig:intervention}
\end{figure}

We evaluate concept-level interventions in the \textbf{unsupervised} setting, where the model has the least prior knowledge of the anomalous domain and thus stands to benefit most from human feedback.
As shown in Fig.~\ref{fig:intervention}, anomaly detection performance improves consistently as more concepts are intervened upon, with most MVTec categories approaching near-perfect AUC once a majority of concepts are corrected; this confirms the value of human-in-the-loop refinement for practical and safety-critical applications, where practitioners can trade off annotation effort against detection accuracy.
\\
Two categories, \textit{transistor} and \textit{toothbrush}, show a degradation: both have few concepts, making the hard label imposed by intervention over-constrain the prediction.
Nevertheless, the overall trend is positive, and this capability is a unique advantage of \ourmethod{} over standard \ac{VAD} methods that, lacking a concept bottleneck, cannot expose any mechanism for structured human correction.

\subsection{Ablation Study}\label{subsec:ablation}
Table~\ref{tab:united_results}~(right) presents an ablation of \ourmethod{} across supervision levels, reporting the contribution of the Concept branch, the Visual branch, and their fusion.
Our Visual branch improves over STFPM at the pixel level: as shown in Table~\ref{tab:comparison}, P-AUC increases from 0.88 (STFPM, MobileNet-v2) to 0.91, thanks to teacher network fine-tuning on the concept prediction task, which injects domain knowledge and yields more informative feature extraction \cite{barusco2025paste}.
With respect to PatchCore, our Visual branch trades a lower P-AUC (0.91 vs.\ 0.96) for a higher I-AUC (0.90 vs.\ 0.86) at less than half the memory, and adds concept-based explanations. Remarkably, the Visual branch not only provides intuitive visual explanations to the end user but also enhances the robustness of the model when the Concept branch fails. Moreover, because the Visual branch of \ourmethod{} is based on the student--teacher paradigm, which has been proven effective for detecting novel data, it further improves robustness against previously unseen anomalies that the Concept branch might miss.
Importantly, \ourmethod{} is not tied to any specific Visual branch architecture: any method that operates on a shared feature extractor can serve as the Visual branch, making the methodology broadly applicable. STFPM was chosen here as a representative example due to its efficiency.

\subsection{Synthetic-to-real Alignment}\label{subsec:sim2real}
The usefulness of generated anomalies hinges on whether their concept-logit representations align with those of real defects. We probe this with 2D t-SNE embeddings of the concept logits on \texttt{metal\_nut}, a visually complex category (Fig.~\ref{fig:tsne_metalnut}). In the unsupervised setting, synthetic anomalies form a tight cluster separated from normal samples, while real test anomalies fall largely \emph{within} the normal region: generated and real defects are misaligned, and concepts transfer poorly (C-AUC $0.59$, I-AUC $0.66$). This is specific to hard categories: in simpler ones such as hazelnut, synthetic and real anomalies overlap and transfer well (see Supplementary Material). Adding a single real anomaly per defect type (semi-supervised) pulls real anomalies toward the synthetic cluster, raising performance to C-AUC $0.70$ and I-AUC $0.84$: even minimal supervision is highly effective at bridging the synthetic-to-real gap.

\definecolor{tnormal}{HTML}{19D3F3}
\definecolor{tanom}{HTML}{EF553B}
\definecolor{testnormal}{HTML}{636EFA}
\definecolor{testanom}{HTML}{FFA15A}
\begin{figure}[thbp]
\centering
\begin{minipage}{0.45\linewidth}
    \centering
    \includegraphics[width=\linewidth]{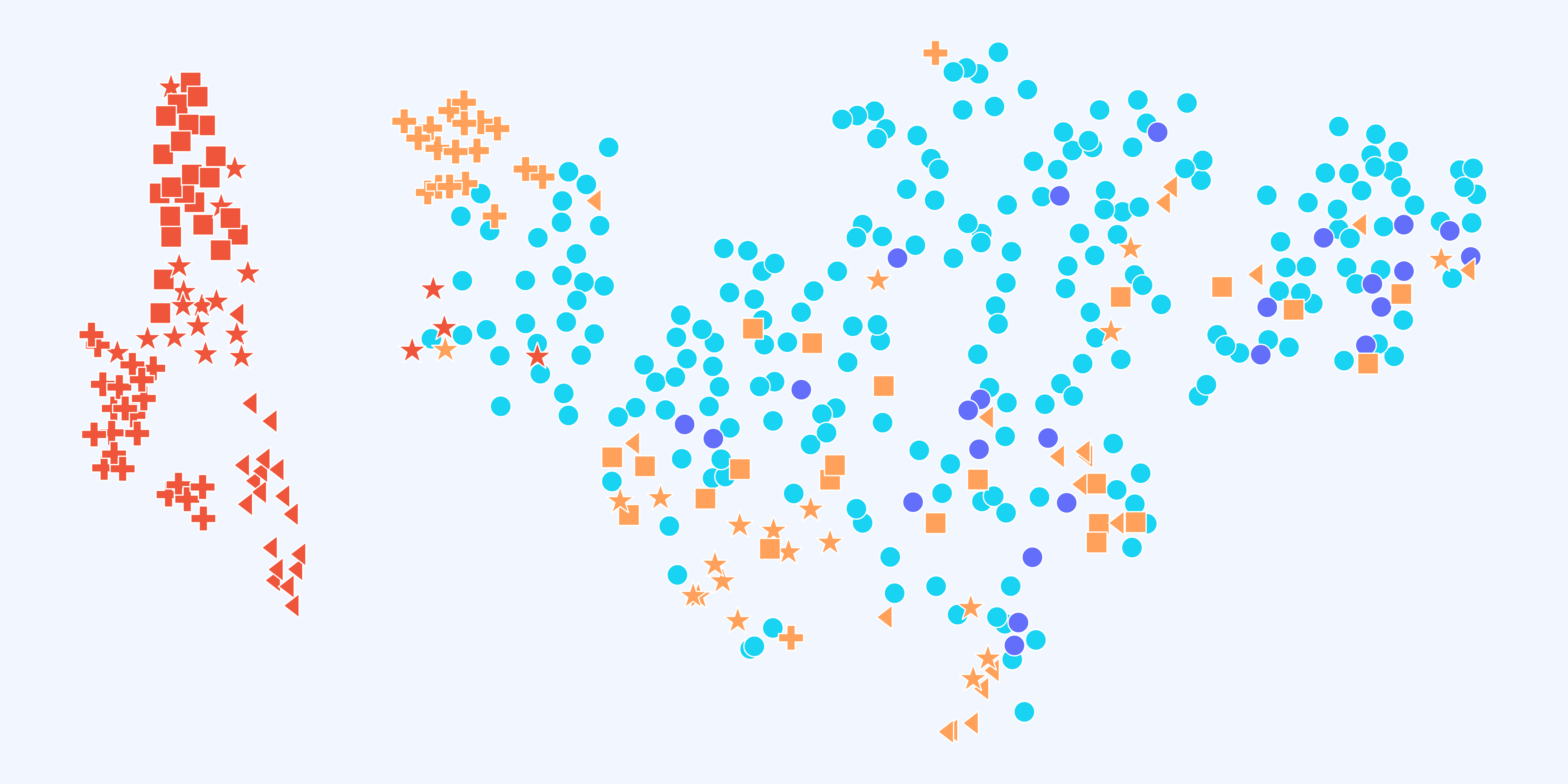}
\end{minipage}\hfill
\begin{minipage}{0.45\linewidth}
    \centering
    \includegraphics[width=\linewidth]{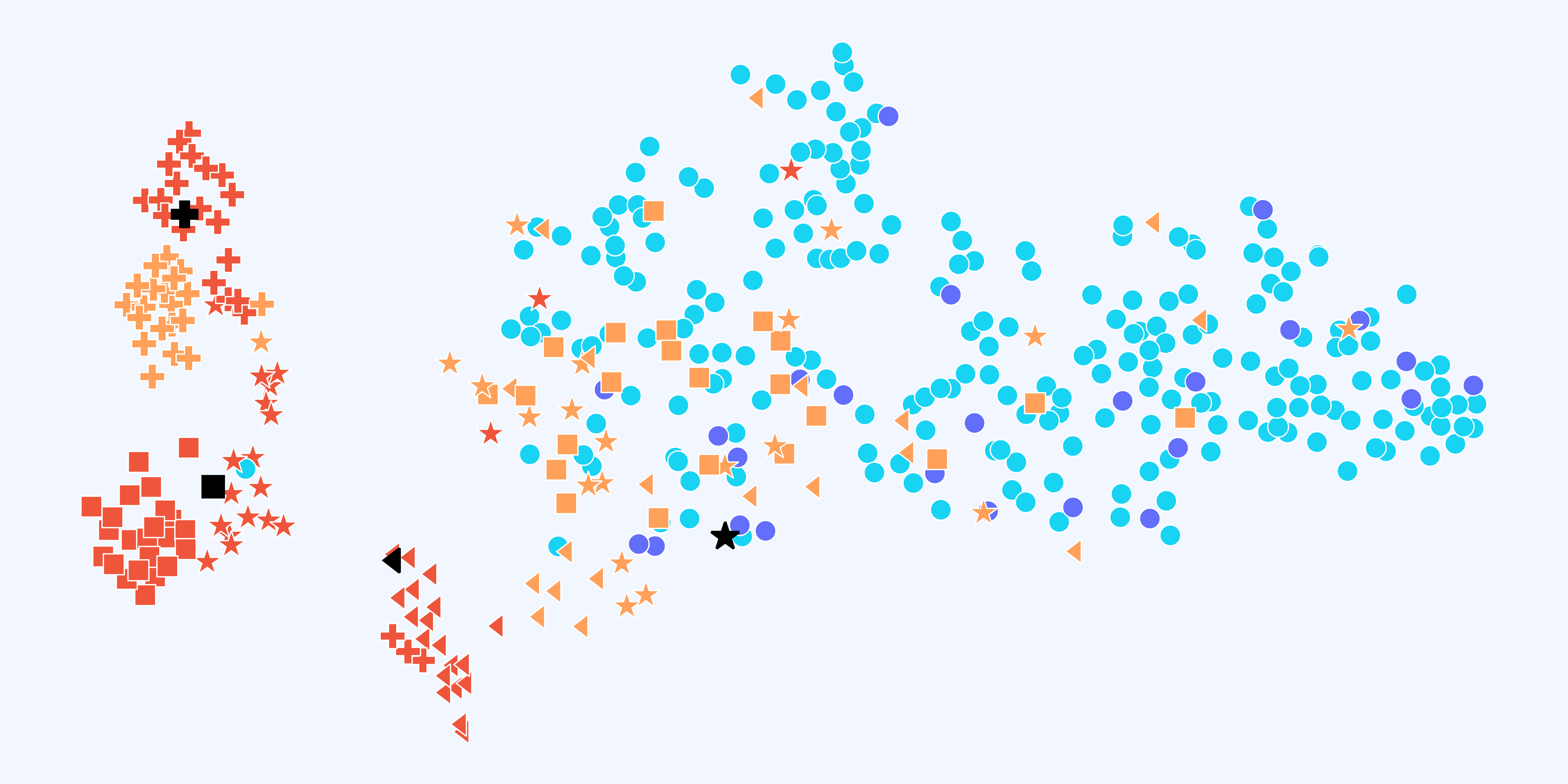}
\end{minipage}
\caption{t-SNE of concept-logit embeddings for \texttt{metal\_nut} in the (Left) unsupervised and (Right) semi-supervised settings. Color encodes the split/label
(\textcolor{tnormal}{$\bullet$}~train normal,
\textcolor{tanom}{$\bullet$}~train anomalous (generated),
\textcolor{testnormal}{$\bullet$}~test normal,
\textcolor{testanom}{$\bullet$}~test anomalous (real),
$\bullet$~train anomalous (real)),
while marker shape encodes the defect type
($\bullet$~good, ${\scriptstyle\blacksquare}$~bent, $\blacktriangleleft$~color, $\pmb{+}$~flip, $\bigstar$~scratch).
Generated anomalies (Left) cluster apart from the real test anomalies; adding one real anomaly per defect type (Right) pulls real and synthetic anomalies into alignment.}%
\label{fig:tsne_metalnut}
\end{figure}

\section{Conclusion}%
\label{sec:conclusion}

This work introduces \ourmethod, adapting \acp{CBM} to \ac{VAD} to bridge high-performing detection and human-understandable reasoning. Pairing a supervised \ac{CBM}-based Concept branch for concept-level explanations with a Visual branch for pixel-level localization, \ourmethod{} delivers dual interpretability that no existing \ac{VAD} method offers.

Across MVTec AD, VisA, and Real-IAD, \ourmethod{} matches state-of-the-art detection while providing transparent, concept-grounded predictions. Concept-level interventions yield substantial gains, confirming the value of the human-machine collaboration the \ac{CBM} formulation enables. Benchmarking supervision regimes shows even an unsupervised fully synthetic-anomalies setting is viable, while combining few real anomalies with synthetic defects is a robust, cost-effective strategy.

The main limitation is the reliance on anomalous training samples: synthetic generation reduces this dependency but does not yet match real anomalies. Future work will improve synthetic generation quality and exploit the Visual branch for novelty detection, capturing unseen defects where the Concept branch alone fails, closing the gap toward fully unsupervised, interpretable and efficient \ac{VAD}. Moreover, we plan on devising a strategy to effectively unify the concept and visual branch so that they process the same signal, without incurring a loss in performance.

\bibliography{main}

\clearpage
\setcounter{page}{1}

\title{Supplementary Material for\\``Explainable Visual Anomaly Detection via Concept Bottleneck Models''}
\maketitle

\section{Supplementary Material}

\section{Prompts Details}
\subsection{Concept Extraction Prompt}
We fabricate the prompt by adding context cues that can guide the model in extracting meaningful concepts: we specify which object is present in the pictures, whether it is anomalous or not and, if so, which defect is present. Next, we employ a Chain-of-Thought prompting strategy and formulate two consecutive prompts:
\begin{enumerate}
    \item \textit{``Provide a description of the image that includes information about all the relevant features that are visible. Focus only on what can be seen, avoiding speculations or assumptions''.}
    \item \textit{``Provided the following description, extract the five most meaningful concepts. Concepts should be defined in such a way that, observing the picture, it is possible to clearly answer with yes or no about its presence''.}
\end{enumerate}
It is worth mentioning that this step is applied to a small subset of the dataset, corresponding to 5\% of it, so it requires the availability of only a few labeled anomalous images. When labeled anomalous samples are not available, the same procedure can be applied to synthetically generated pictures, keeping in mind that the quality of the extracted concepts is bounded by the quality of the generation process.

\subsection{Dataset Annotation Prompt}
A very similar prompt for dataset annotation as we did for concept extraction is employed: for each image, we include information about the object and, if present, the type of anomaly.
\\
\textit{``I provide an image of a \{category\}. The image has been classified as \{label\}, [which implies that it shows a visible defect or anomaly, specifically \{defect type\}.] Knowing this, choose among the following list of concepts which ones can be clearly seen in the picture. Output the result as a JSON object of the following form: \{Concept 1: True, Concept 2: False, \ldots \}''.}
\\
Through the previous prompt, we ensure to focus only on binary concepts.

\subsection{Anomaly Generation Prompt}
\begin{figure}
\centering
\includegraphics[width=0.9\linewidth]{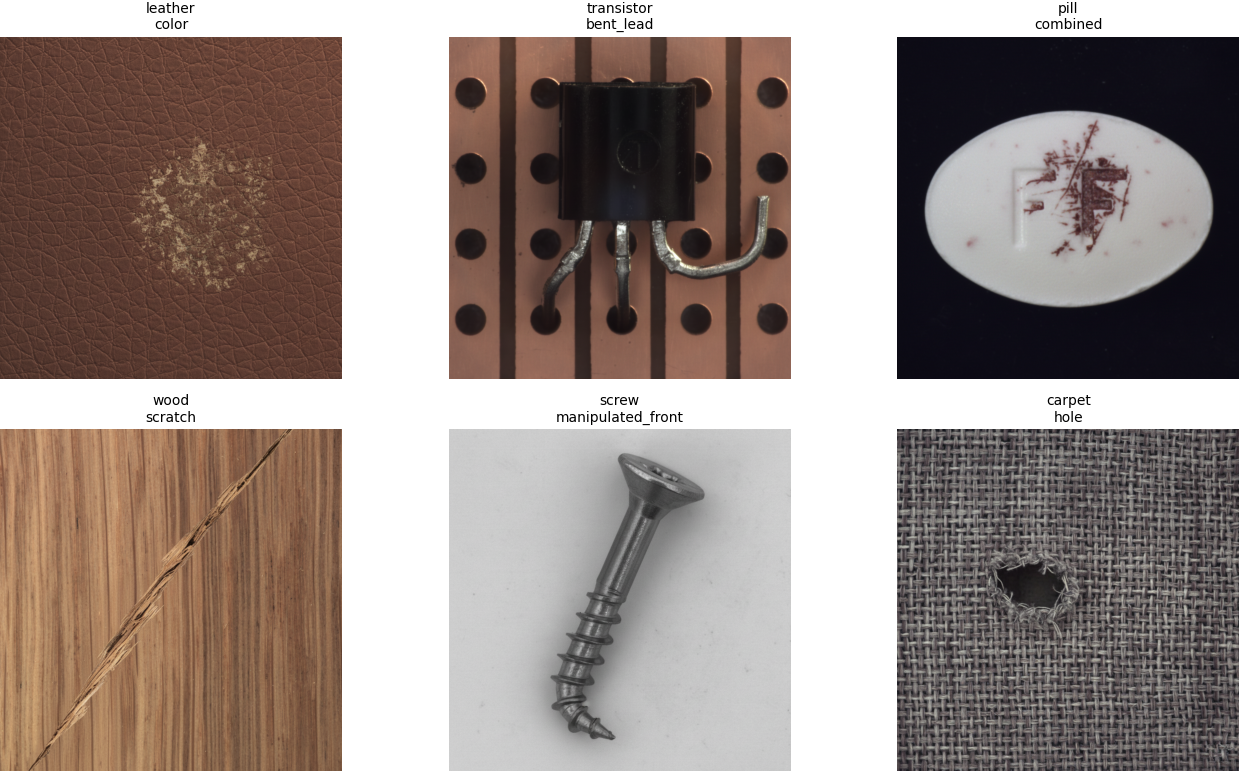}
\caption{Examples of well-generated synthetic anomalous images.}%
\label{fig:gen_anomalies}
\end{figure}

We generate anomalous images leveraging the capabilities of Google's Nano Banana. We assume we have a set of normal pictures and a list of defects that we want to include in the synthetic images, provided by a domain expert. Next, we query a \ac{LLM}, in our case GPT-5, to provide ten synonyms of the defect we want to add, to ensure richer variability, and build prompts according to this structure:
\\
\textit{``Modify this image by adding a \{type of defect\} to the \{name of the object\}, keeping the same angle, view, and pose.''}
\\
Some categories and defect types, which proved to be particularly challenging for the \ac{VLM}, required manually tuning the previous prompt and discarding some of the generated images. Some examples of the synthetic pictures obtained can be found in Fig.~\ref{fig:gen_anomalies}.

\section{Evaluation of the Concept Annotation Pipeline}
In addition to the metrics reported in the main body (91.3\% accuracy achieved by the automated annotation pipeline against a subset of 400 manually-annotated images), we  also evaluate the full automatically annotated dataset of the \textit{hazelnut} category against a ground-truth version of it, also computing precision and recall to make a more detailed analysis. Table~\ref{tab:concept_accuracy} displays a summary of our findings. Overall, predicted concepts can be considered of good quality, especially those that appear more often in normal images; as for anomalous concepts, we observe a very high variability in terms of precision and recall, which is mainly related to those concepts that describe a diffused feature of the image (e.g., \textit{surface discontinuity, uneven tone, etc.}), which are more difficult to capture for a \ac{VLM} and can be arbitrary for a domain expert as well.

\begin{table}[h]
\centering
\renewcommand{\arraystretch}{1.2}
\setlength{\tabcolsep}{3pt}

\begin{tabular}{l c c c}
\toprule
\textbf{} & \textbf{All} & \textbf{Anomaly Concepts} & \textbf{Normal Concepts} \\
\midrule
\textbf{Acc}    & 0.93 $\pm$ 0.08 & 0.92 $\pm$ 0.08 & 0.98 $\pm$ 0.01 \\
\textbf{Prec}   & 0.76 $\pm$ 0.24 & 0.71 $\pm$ 0.23 & 0.99 $\pm$ 0.00 \\
\textbf{Rec}    & 0.77 $\pm$ 0.28 & 0.72 $\pm$ 0.28 & 0.99 $\pm$ 0.01 \\
\bottomrule
\end{tabular}
\caption{Average value and standard deviation of accuracy, precision and recall of predicted concepts over the \textit{hazelnut} category.}%
\label{tab:concept_accuracy}
\end{table}

\section{Additional Training Details}

Each model is trained for a maximum of 100 epochs, with an early stopping mechanism triggered after ten epochs. When performance reaches a plateau after five epochs, we also allow the learning rate to be reduced by a factor of 10.
\\
\textbf{Transformation-based Data Augmentation.}
We apply transformations to enhance the model generalization ability, carefully choosing those that preserve the presence of each concept, ensuring the integrity of attribute labels.
The transformations applied are the following: horizontal and vertical flips with probability 0.5, random rotation by a 25-degree angle, brightness jitter by a factor of 0.2, and contrast jitter by a factor of 0.2.
\\
\textbf{CBM Training.}
We attach $k$ parallel linear layers to the CNN feature extractor that act as concept predictors, while a simple Feed-Forward Neural Network with eight neurons is employed for anomaly detection. The feature extractor undergoes a pre-training phase directly on the dataset of interest, in which it is optimized according to a multi-class classification objective to learn which object is depicted in the image, while its last layers are fine-tuned during training for the concept prediction task.

\section{CBM training Modalities}
Together with the joint paradigm described in the main paper, a \ac{CBM} can be trained using two alternative modalities.

\smallskip
\noindent \textbf{Independent training}:
          the concept extractor is trained to predict concepts from inputs:
          \begin{equation}
              \hat{g} = \arg \min_g \sum_{i,j} \mathcal{L}_{C_j}(g_j(\mathbf{x}^i), c_j^i),
              \label{eq:indtrain}
          \end{equation}
          where $\mathcal{L}_{C_j}$ denotes the loss associated with the $j$-th concept prediction.
          Separately, the label predictor is trained using the ground-truth concept vectors $\mathbf{c}$ as inputs.
          \begin{equation}
              \hat{f} = \arg \min_f \sum_i \mathcal{L}_Y(f(\mathbf{c}^i), y^i)
          \end{equation}
          where $\mathcal{L}_Y$ is the loss for the downstream label prediction.
          Note that, during deployment, $\hat{f}$ takes as input the predicted concept vector $\mathbf{\hat{c}} = \hat{g}(\mathbf{x})$ rather than the true ones. This may lead to a shift between the training and inference input distributions, potentially degrading performance.

\smallskip
\noindent \textbf{Sequential training}: first, $g$ is trained as in the independent setting; 
    then, $f$ is trained directly on these predicted concepts $\hat{\mathbf{c}} = \hat{g}(\mathbf{x})$ rather than the ground truth ones.

Table~\ref{tab:results_cbm_paradigms} displays a comparison of the results obtained by the three \ac{CBM} paradigms on MVTec AD.
In consideration of these findings, the \ac{CBM} model was trained using the Joint paradigm, since it performs similar to the Independent approach, but with a slightly better I-AUC.

\begin{table}
    \centering
    \renewcommand{\arraystretch}{1.2}
    \setlength{\tabcolsep}{3pt}

    \begin{tabular}{l c c}
        \toprule
        \textbf{Model type}  & \textbf{C-AUC} & \textbf{I-AUC} \\
        \midrule
        \textbf{Joint}       & 0.86           & 0.97           \\
        \textbf{Sequential}  & 0.85           & 0.95           \\
        \textbf{Independent} & 0.87           & 0.96           \\
        \bottomrule
    \end{tabular}
    \caption{Comparison of the results of the three \ac{CBM} learning paradigms over the MVTec AD dataset, in terms of concept prediction performance and anomaly detection.}%
    \label{tab:results_cbm_paradigms}
\end{table}

\section{Ablation Studies on Fusion and Calibration Strategies}
We conducted an ablation study on MVTec AD (unsupervised setting) analyzing (i) score calibration for the visual and concept branches and (ii) score fusion. Results are reported in Table~\ref{tab:ablation_fusion}, including agreement with the concept branch (\emph{agree}), i.e., the percentage of samples whose combined-score binary prediction matches the concept branch.
\begin{enumerate}[label=(\roman*), wide=0pt, itemsep=0.5em]
    \item \textbf{Calibration:} We normalize visual and concept anomaly scores into percentile ranks of their empirical test distributions, mapping both to a common scale without supervision. Compared with supervised isotonic regression (trained on a small held-out set), rank-based normalization consistently performs better. 
    \item \textbf{Fusion:} We compare the proposed max operator against the arithmetic mean, a logistic regression on the two calibrated scores, and a logistic regression that additionally uses concept probabilities. Although the learned fusion strategies achieve higher AUROC, they consistently reduce agreement with the concept branch. Since explanations are only faithful to the concept branch, we retain the max operator and leave improved fusion strategies for future work.
\end{enumerate}

\begin{table}[t]
    \centering
    \footnotesize
    \setlength{\tabcolsep}{5pt}
    \renewcommand{\arraystretch}{0.88}
    \resizebox{0.65\columnwidth}{!}{%
    \begin{tabular}{l c c c c}
        \toprule
        & \multicolumn{2}{c}{\textbf{Rank}} & \multicolumn{2}{c}{\textbf{Isotonic Reg.}} \\
        \cmidrule(r){2-3} \cmidrule(l){4-5}
        \textbf{Fusion} & \textbf{AUROC} & \textbf{Agree} & \textbf{AUROC} & \textbf{Agree} \\
        \midrule
        \textbf{Max}       & 0.81          & \textbf{87.31\%} & 0.76 & 84.62\% \\
        \textbf{Mean}      & 0.88          & 82.95\%          & 0.87 & 73.29\% \\
        \textbf{Learned}   & 0.90          & 79.49\%          & 0.87 & 73.28\% \\
        \textbf{Learned+C} & \textbf{0.91} & 76.52\%          & 0.88 & 78.50\% \\
        \bottomrule
    \end{tabular}%
    }
    \caption{Ablation study on calibration (columns) and fusion (rows) strategies in the unsupervised setting on MVTec AD.}
    \label{tab:ablation_fusion}
\end{table}

\section{Additional User Study Details}
Participants (32 grad students/researchers, no domain expertise required) received samples via a deterministic per-participant seed with inverse-frequency sampling balanced across method, dataset, category, and label, yielding mostly disjoint stimuli across methods (7.7\% image reuse) and each method's natural correct/incorrect ratio (unenforced). CONVAD's helpfulness advantage over AnomalyGPT is significant ($\chi^2$, $p{<}0.001$; participant-cluster bootstrap 95\% CI excludes zero), while it is on par with Anomaly-OV ($p{=}0.36$). We acknowledge two limitations: heatmaps are shown only for CONVAD/AnomalyGPT (Anomaly-OV produces none), and stimuli are not matched across methods.

\section{Intervention Procedure}
In Sec.~\ref{subsec:intervention}, we demonstrate the impact on performance of manually modifying predicted concepts with their ground-truth value during inference. However, this procedure can be costly, as it requires the supervision of a human expert, so several strategies have been devised to minimize such costs by focusing first on the concepts that should be more important to provide an increase in performance.
We followed the UCP (Uncertain Concept Prediction) heuristic proposed in~\cite{shin2023closer} and computed the entropy-based uncertainty related to each concept prediction:
\begin{equation}
    \mathcal{H}(c_i)=-(p_i \cdot \text{log}(p_i)+(1-p_i)\cdot \text{log}(1-p_i)).
\end{equation}
We then sorted the entropy scores in descending order, following the idea that concepts predicted with more uncertainty might confuse the model and lead to incorrect predictions. As mentioned in~\ref{sec:cbm}, since the joint model uses the predicted concept logits to perform the main task, we substitute them with either the 5th or the 95th percentile of the training distribution.

\section{Concept Logits Analysis}
In the unsupervised setting, where the model is trained exclusively on synthetic anomalies and evaluated on real ones, some categories achieve high concept-level AUC (C-AUC) and image-level AUC (I-AUC), while others fail to transfer effectively to real data. Two extreme examples illustrating this contrast are hazelnut, which transfers well, and metal nut, which does not.
We examine the concept-logit embeddings extracted from the training and test examples to investigate this discrepancy. The test set contains only real normal and real anomalous examples, while the training set may include synthetic images in the SAG scenarios.
All models compared within each category use the same random seed, training data, and evaluation set, with the latter containing only real anomalies that were not used for weak supervision in any of the models.

A well-learned bottleneck should ensure (i) a separation between normal and anomalous samples, provided that the concept set is sufficiently predictive, and (ii) tight clustering of images sharing the same or similar defect type (\eg ``hole'' or ``crack'' in hazelnut). Moreover, if the synthetic anomalies are sufficiently well aligned with the real anomaly domain, their concept-logit representations should be consistent with those of real anomalies, indicating an effective transfer. To obtain qualitative insights about these aspects, we plot 2-dimensional t-SNE visualizations of the concept logits.

\begin{figure}
\centering
\includegraphics[width=0.6\linewidth]{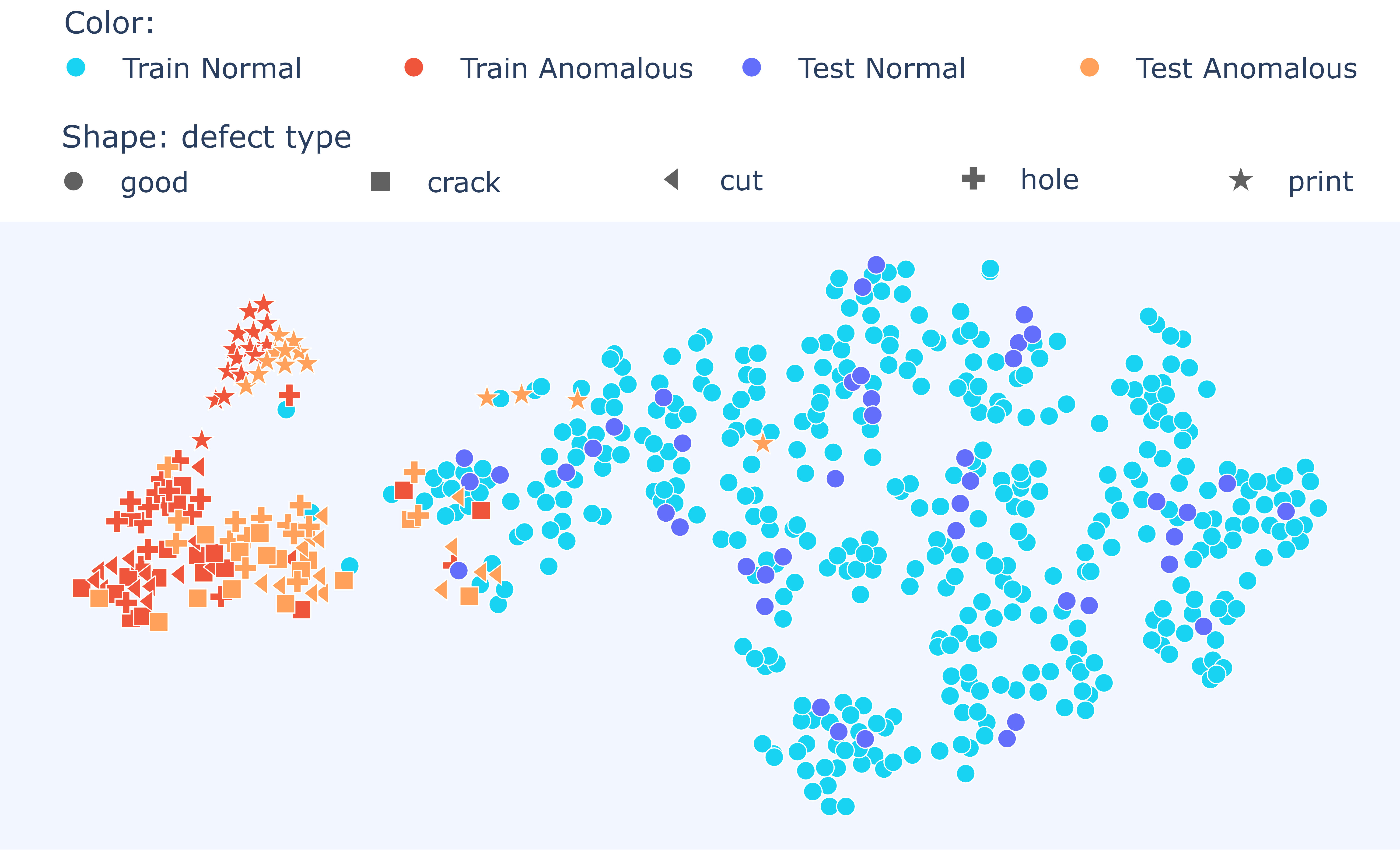}
\caption{Hazelnut t-SNE embeddings for the SAG scenario.}%
\label{fig:conceptlogits_hazelnutSAG}
\end{figure}

In the hazelnut SAG scenario, we obtain an average C-AUC of 0.89 and an I-AUC of 0.99. Fig.~\ref{fig:conceptlogits_hazelnutSAG} shows a clear separation between \texttt{Train Normal} and \texttt{Train Anomalous} (synthetic) samples.
\texttt{Test Normal} samples appear in the same region as \texttt{Train Normal}, which is expected since their domains match, as both correspond to real images.
In addition, \texttt{Test Anomalous} (real images) are mapped to the same broad regions as synthetic anomalies (\texttt{Train Anomalous}), which is ideal.
Within this anomalous region, the ``print'' defect forms a distinct sub-cluster, while ``crack'', ``cut'' and ``hole'' share more overlapping embeddings, reflecting their visual similarity.
These patterns are consistent with the high AUC scores and indicate that the \textit{synthetic anomalies generated for the hazelnut category effectively capture relevant characteristics of real anomalies}.

\begin{figure}
\centering

\begin{minipage}{0.48\linewidth}
    \centering
    \includegraphics[width=\linewidth]{images/supp/tsne_metalnut_SAG.png}\\
    (a) SAG scenario
\end{minipage}
\hfill
\begin{minipage}{0.48\linewidth}
    \centering
    \includegraphics[width=\linewidth]{images/supp/tsne_metalnut_W1SAG.png}\\
    (b) Supervised (1 anomaly-shot-per-defect)+SAG scenario
\end{minipage}

\vspace{0.5em}

\begin{minipage}{0.48\linewidth}
    \centering
    \includegraphics[width=\linewidth]{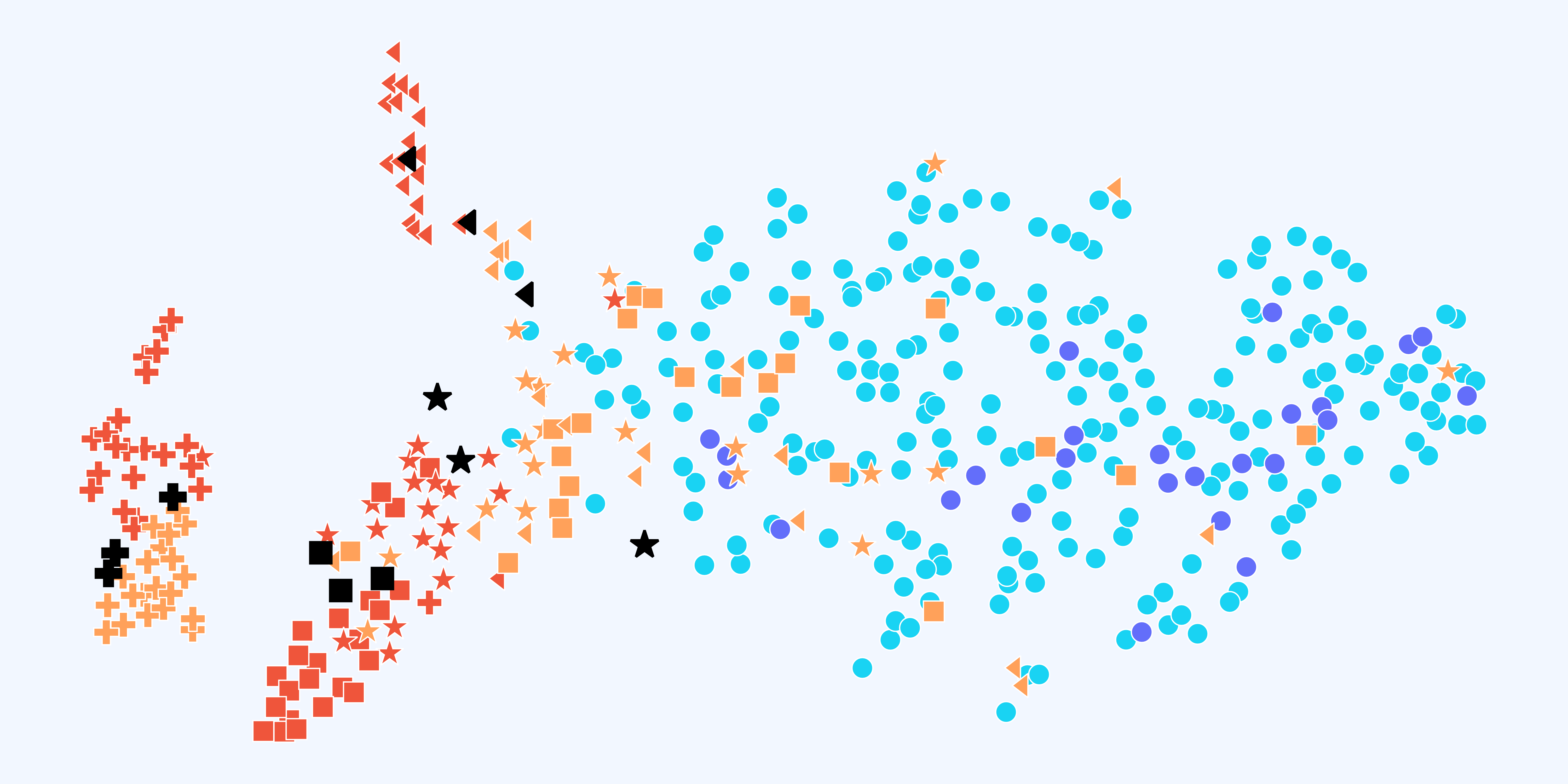}\\
    (c) Supervised (3 anomaly-shot-per-defect)+SAG scenario
\end{minipage}
\hfill
\begin{minipage}{0.48\linewidth}
    \centering
    \includegraphics[width=\linewidth]{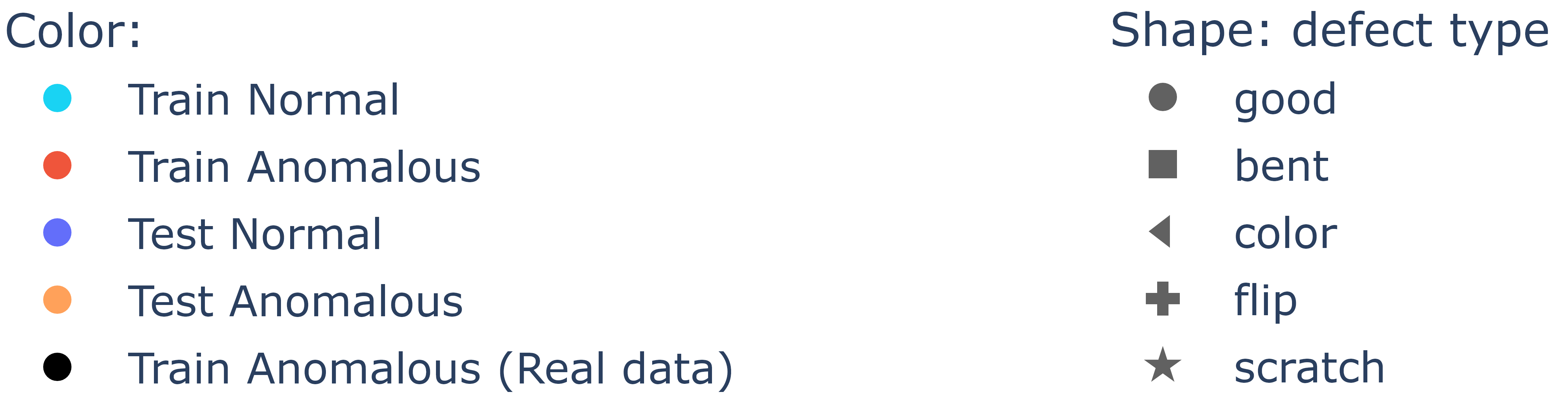}
\end{minipage}

\caption{t-SNE representations of concept logits embeddings using \texttt{metal\_nut}. Black markers highlight the few real samples used for Supervised+SAG scenarios.}%
\label{fig:2x2plot}
\end{figure}

However, for the metal nut in the SAG scenario, the C-AUC drops to an average of 0.59, and the I-AUC drops to 0.66.
In this case, in Fig.~\ref{fig:2x2plot}(a), most \texttt{Test Anomalous} (real) samples are located within regions corresponding to \texttt{Train Normal} and \texttt{Test Normal} data, rather than being mapped near the synthetic \texttt{Train Anomalous} samples.
Only the ``flip'' defect forms a slightly separated cluster (due to its visual dissimilarity to other images), while other defects are scattered among normal samples. This qualitative mismatch between synthetic and real anomaly embeddings explains the degraded AUCs and suggests poorly generated anomalies.

Exploring the embeddings in the Supervised+SAG scenarios provides a deeper understanding of why introducing even a small number of real anomalies, combined with synthetic anomalies, reduces this discrepancy in performance.
For the metal nut in the Supervised (1 anomaly-shot-per-defect)+SAG scenario (adding one real anomaly per defect type in training), the C-AUC improves to an average of 0.70 and the I-AUC to 0.84, and Fig.~\ref{fig:2x2plot}(b) shows a closer alignment between synthetic and real anomalies.
With Supervised (3 anomaly-shot-per-defect)+SAG (three real anomalies per defect type, compared to one in the standard semi-supervised scenario), the C-AUC increases to 0.76 and the I-AUC to 0.89. Simultaneously, the representations of real anomalies appear closer to those of synthetic ones, and clusters associated with different defect types become more clearly separated (Fig.~\ref{fig:2x2plot}(c)).
These results indicate that when using synthetic anomalies for training, even minimal real-data supervision can mitigate the domain shift and encourage the concept space to encode a more consistent defect-specific structure, eliminating the need to collect a large dataset of rare and difficult-to-acquire anomalies.

\section{Additional Concept Vocabulary Analysis}

\begin{figure}[h]
\centering
\includegraphics[width=0.92\linewidth]{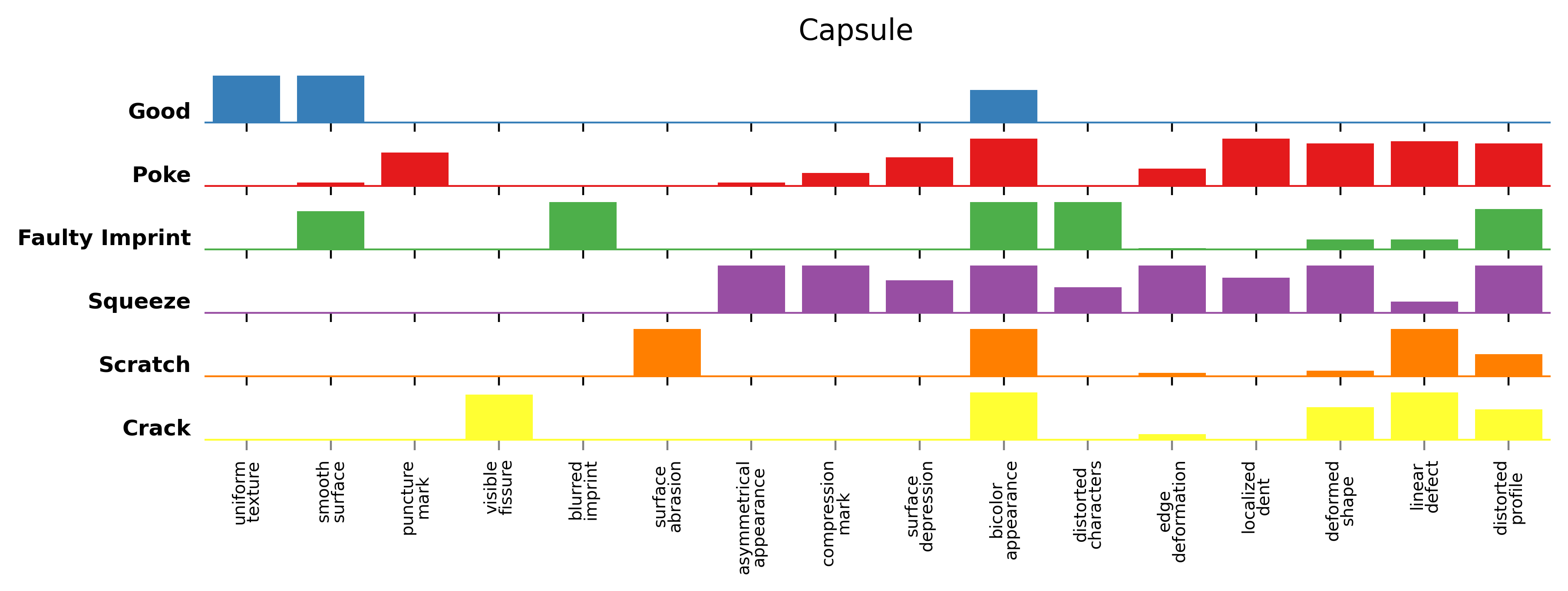}
\caption{Concept distribution for the \texttt{capsule} category (MVTec AD).}%
\label{fig:concept_dist_capsule}
\end{figure}

\begin{figure}[h]
\centering
\includegraphics[width=0.92\linewidth]{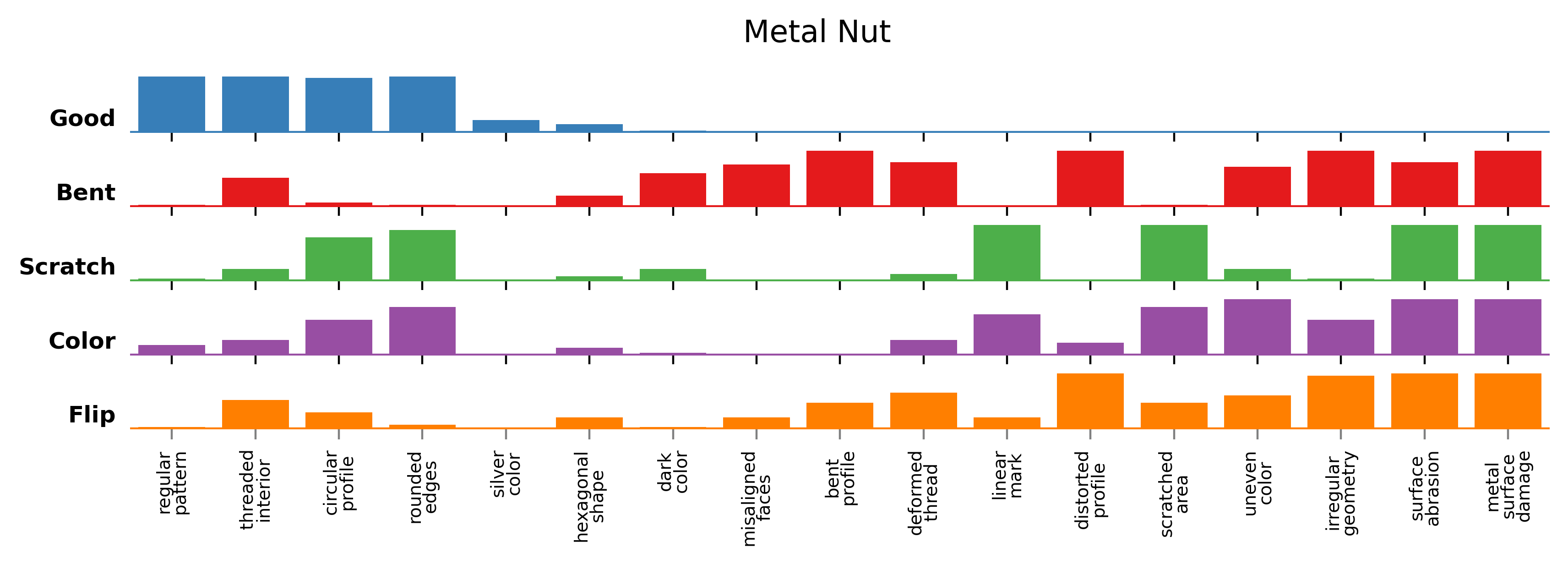}
\caption{Concept distribution for the \texttt{metal\_nut} category (MVTec AD).}%
\label{fig:concept_dist_metal_nut}
\end{figure}

\begin{figure}
\centering
\includegraphics[width=0.92\linewidth]{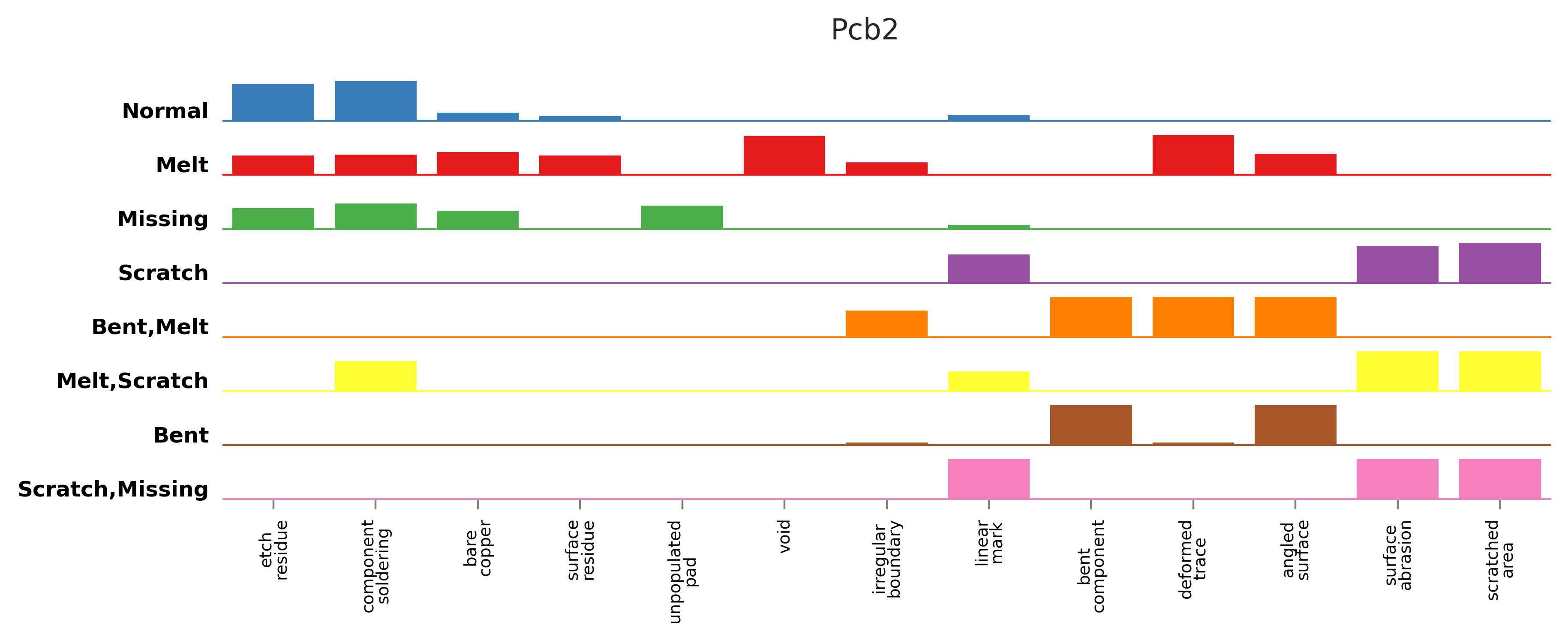}
\caption{Concept distribution for the \texttt{pcb2} category (VisA).}%
\label{fig:concept_dist_pcb2}
\end{figure}

\begin{figure}
\centering
\includegraphics[width=0.92\linewidth]{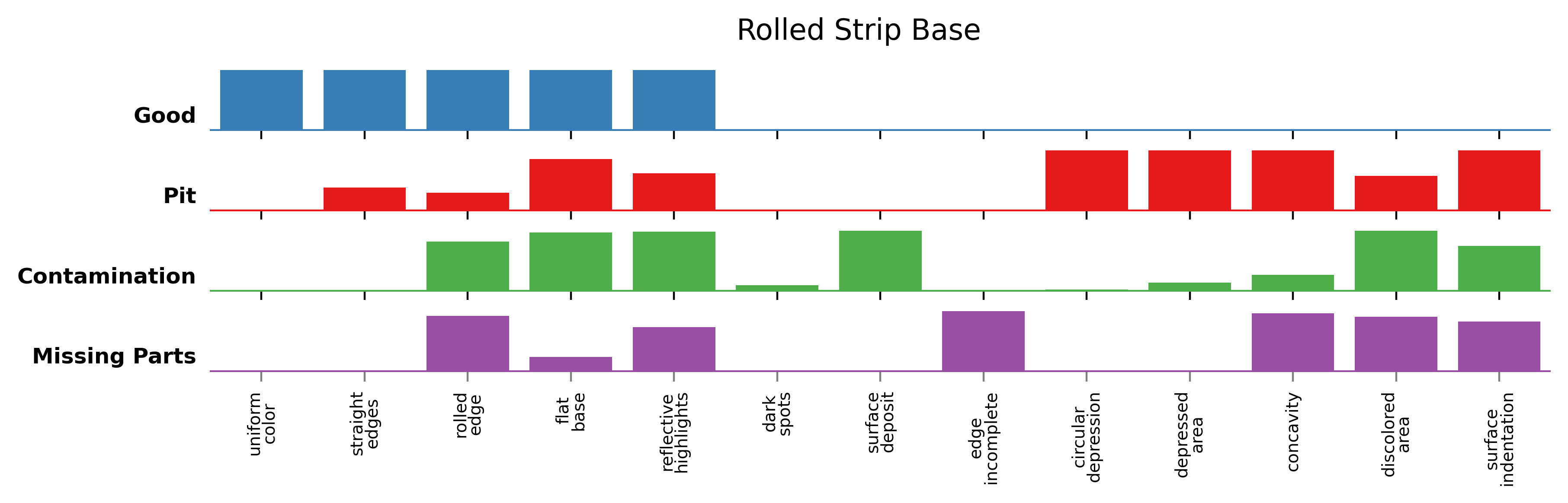}
\caption{Concept distribution for the \texttt{rolled\_strip\_base} category (Real-IAD).}%
\label{fig:concept_dist_rolled_strip}
\end{figure}

We report the concept distributions for representative object categories from each of the three benchmarks (Figs.~\ref{fig:concept_dist_capsule}--\ref{fig:concept_dist_rolled_strip}) to illustrate how the extracted concepts capture meaningful characteristics of the data. In particular, the distributions highlight clear differences between normal and anomalous samples, showing that concepts activated for specific defect types can help identify the nature of the fault from an interpretability standpoint.

\end{document}